\documentclass[letterpaper]{article}
\usepackage{emnlp2021}
\usepackage{times}
\usepackage{helvet}
\usepackage{courier}
\usepackage{graphicx}
\usepackage{booktabs}
\usepackage{xcolor}
\usepackage{xspace}
\usepackage{graphicx}
\usepackage{subfig}
\usepackage{multirow}
\usepackage{array}
\usepackage{verbatim}
\usepackage{mdwlist}

\usepackage[english]{babel}
\usepackage{blindtext}

\usepackage[utf8]{inputenc} %
\usepackage[T1]{fontenc} %

\usepackage{url}

\usepackage{xr} 
\makeatletter
\newcommand*{\addFileDependency}[1]{%
  \typeout{(#1)}
  \@addtofilelist{#1}
  \IfFileExists{#1}{}{\typeout{No file #1.}}
}
\makeatother
\newcommand*{\myexternaldocument}[1]{%
    \externaldocument{#1}%
    \addFileDependency{#1.tex}%
    \addFileDependency{#1.aux}%
}
\myexternaldocument{supplementalmaterial}

\newcommand{\myparagraph}[1]{\paragraph{#1}}

\newcommand{\so}{{{\textsc{SigOther}}}\xspace}
\newcommand{\pers}{{{\textsc{Person}}}\xspace}

\newcommand{\Sref}[1]{\S\ref{#1}}

\newcommand{\sref}[1]{\S\ref{#1}}
\newcommand{\fref}[1]{Figure~\ref{#1}}

\title{Using Sociolinguistic Variables to Reveal Changing Attitudes \\Towards Sexuality and Gender}

 \author{
    Sky CH-Wang$^\dagger$ \\
    Department of Computer Science \\
    Columbia University \\
    \texttt{skywang@cs.columbia.edu}\\ 
    \And
    David Jurgens \\
    School of Information \\
    University of Michitgan \\
    \texttt{jurgens@umich.edu}\\}
   
\date{}

\begin{document}
\maketitle
\begin{abstract}

Individuals signal aspects of their identity and beliefs through linguistic choices. Studying these choices in aggregate allows us to examine large-scale attitude shifts within a population. 
Here, we develop computational methods to study word choice within a sociolinguistic lexical variable---alternate words used to express the same concept---in order to test for change in the United States towards sexuality and gender. We examine two variables: i) referents to significant others, such as the word ``partner'' and ii) referents to an indefinite person, both of which could optionally be marked with gender. The linguistic choices in each variable allow us to study increased rates of acceptances of gay marriage and gender equality, respectively. In longitudinal analyses across Twitter and Reddit over 87M messages, we demonstrate that attitudes are changing but that these changes are driven by specific demographics within the United States. Further, in a quasi-causal analysis, we show that passages of Marriage Equality Acts in different states are drivers of linguistic change.

\end{abstract}

\newcommand\blfootnote[1]{%
  \begingroup
  \renewcommand\thefootnote{}\footnote{#1}%
  \addtocounter{footnote}{-1}%
  \endgroup
}
\blfootnote{$^\dagger$Work performed while the author was an undergraduate research assistant at the University of Michigan.}

\section{Introduction}

A person's identity and attitudes are reflected in the language they use \cite{norton1997language,huffaker2005gender,de2012discourse}.
In particular, the linguistic choice for a concept can reveal the individual's stance or attitudes \cite{jaffe2009stance}; for example, the use of ``illegals'' or ``undocumented'' in reference to immigrants reveals the speaker's attitude on immigration \cite{lakoff2006framing}.
These alternations in word choice are known as \textit{lexical variables} in sociolinguistics. 
Examining the relative frequencies of a variable's words  can reveal the underlying attitudes within a population that drive the linguistic choice.
Here, we examine changes in attitude towards sexuality and gender in the United States through two lexical variables.

Sociolinguistics has long focused on variation in language with respect to identity and attitudes \cite{labov1963social,eckert2001style,trudgill2002sociolinguistic}.
Recent computational studies have built upon this line of research \cite{nguyen2016computational}, showing not only that this variation occurs in social media \cite{eisenstein2014diffusion,hovy2015user}, but also that the large scale of social media enables the study of broader societal trends \citep{abitbol2018socioeconomic,grieve2018mapping}. 
Our work expands this line of research by examining longitudinal changes in linguistic variation to show changing societal attitudes.

Here, we test for change in attitudes about sexuality and gender by computationally measuring variation for two lexical variables associated with these attitudes from a massive longitudinal study of 73M Twitter posts and 14M Reddit comments across nearly ten years. The first variable focuses on the use of gender when referring to romantic partners; specifically, we test how frequencies in gender-neutral referents such as \textit{partner}---a term often used by LGBT+ community members---shift as acceptance of gay marriage changes. The second variable measures attitudes about gender through testing for unnecessary gender markings on indefinite references to one or more people, e.g., ``some folks'' versus ``some guys.''
Our work here is drawn from theory in gender and sexuality studies on how both heterosexuality and masculinity are treated as the default or norm in English \cite{kitzinger2005heteronormativity,land2005speaking}, where shifts away from these heterosexual constructs signal increasing acceptance of other identities.

Our paper offers the following three contributions.
First, through a large-scale computational analysis that measures the language choices of different demographics, we demonstrate increasing acceptance of non-heterosexual relationships through the increasing use of non-gendered referents to significant others by heterosexual communities. While non-gendered referents are used frequently in LGBT+ communities, further demographic analysis shows this change is found across gender identities. %
Second, in a quasi-causal analysis, we show that passages of  marriage equality acts (MEA) in the United States drives a statistically significant increase of gendered markers in the LGBT+ community (e.g., \textit{husband} instead of \textit{partner}), mirroring increased acceptance and decreased social cost for explicitly indicating one's sexual orientation \cite{ofosu2019same}.
Third, we find increasing gender equality through decreased use of gendered person referents, driven by multiple segments of the population. %
Our work not only reveals positive societal change in acceptance but points to the potential of linguistic variation as indicator variables for studying cultural attitudes.

\section{Sociolinguistic Variables}

Sociolinguistic variables consist of alternative expressions where each expression is associated with a specific identity or attitude \cite{bucholtz2005identity,eckert2008variation}. 
Typically, these variables have been pronunciations, e.g., the association of g-dropping with African American Vernacular English \cite{wolfram1969sociolinguistic,dillard1973black}, due to the need for high observational frequency within in-person studies in order to  identify rigorous associations between form and identity/stance \cite{labov1972some,labov1981field}. 
The availability of massive quantities of natural text from social media has substantially increased our ability to study lexical variables, which occur less frequently than pronunciation variations  \cite{androutsopoulos2006introduction,nguyen2016computational}. While many studies have focused on associations between demographics and lexical signals \cite{jackson1993early,o2010mixture,jurgens2017writer}, we examine associations between attitudes and two variables: (1) referents to significant others and (2) indefinite referents to one or more persons. We refer to these variables respectively as \so and \pers and motivate them next. %

\myparagraph{\so}
Individuals frequently refer to romantic partners in conversation. Signalling the gender of these partners also reveals their sexual orientation \cite{kitzinger2005speaking,kitzinger2005heteronormativity,wilkinson2015sexuality}. However, in some social contexts, revealing one's non-heterosexual sexuality (i.e., ``outing'') carries social cost and personal risk \cite{fuss2013inside,cadieux2015you,carrasco2018queer}. As a result, some members of the LGBT+ community have adopted gender neutral terms to refer to significant others \cite{whypartner}, e.g., \textit{partner}, as opposed to gendered terms such as \textit{girlfriend} or \textit{husband}. 
Use of the gender-neutral forms is partially predicated on social acceptance of non-heterosexual orientation \cite{land2007iii}; in social settings of acceptance, LGBT+ individuals will readily use and prefer gendered markers for their significant others \cite{heisterkamp2016challenging}.
At the same time, the use of gender-neutral \so terms by LGBT+ community members carries the risk of revealing orientation if the terms are exclusively used by that community. Therefore a concerted effort has been made to adopt gender-neutral terms more broadly so as to decrease their association with sexual orientation \cite{de2018lgbt}.\footnote{However, as a marker of in-community status, this adoption by individuals outside the community has met some resistance \citep{romack_2018,werder_lee_sulc_howse_2017}}
Given the association between social acceptance and linguistic choice within \so (Table \ref{tab:variables}, top), we expect that changing attitudes should result in a change in linguistic behavior.

\begin{table}[t]
    \centering
    \begin{tabular}{l}
    \hline
        \so: \underline{boyfriend}, \underline{\smash{girlfriend}}, \underline{\smash{husband}}, bae,\\ partner, \underline{\smash{bf}}, \underline{\smash{gf}}, babe, \underline{\smash{lover}} \\ [1.2ex] 
        \pers: people, \underline{\smash{girl}}, \underline{\smash{man}}, \underline{\smash{guy}}, person, \underline{\smash{girls}}, \\ \underline{\smash{guys}}, \underline{\smash{dude}}, \underline{\smash{bro}}, individual\\
        \hline
    \end{tabular}
    \caption{Top 10 terms used in the \so and \pers lexical variables; variants are shown sorted by frequency. Gendered markers are underlined.}
    \label{tab:variables}
\end{table}

\myparagraph{\pers}
In the late 20th century, English has seen a shift away from using masculine forms to refer to mixed or other gender individuals or groups \cite{foertsch1997search,earp2012extinction}, e.g., ``you guys'' to refer to a group of any gender. This shift has included increasing use of ungendered pronouns, e.g,. \textit{they} to refer to a single person \cite{lascotte2016singular} and a move away from assuming a particular pronoun \cite{balhorn2004rise}.
In certain contexts, individuals make indefinite references to people or groups, e.g., describing hypothetical examples or evoking a generic use of the term. These settings also allow for gendered and ungendered referents, such as \textit{guys} versus \textit{folks}. Their linguistic choices in these circumstances reflect unconscious biases about a default gender, perpetuating hegemonic masculinity attitudes \cite{cooper2002boys}.
Therefore, in studying the variation in gender marking of indefinite references, we expect that decreases in explicit gendering would coincide with shifts in attitudes towards gender equality and hegemonic masculinity. 

\myparagraph{Measuring Linguistic Choice}
We measure changes in linguistic behavior by fitting a bigram language model $p(w_i|w_{i-1})$ and comparing relative probabilities for each variables' words in restricted contexts.
The words comprising each variable were identified through a review of past literature \citep{kiesling2004dude,heisterkamp2016challenging} and inclusion of unambiguous synonyms. Table \ref{tab:variables} lists the most common variants, with the rest in supplemental material.
Context restriction is necessary as not all uses of the words in our variables correspond to the sense intended for study, e.g., the use of \textit{partner} in  ``business partner'' does not reflect a choice within \so. Therefore, we apply a set of syntactic heuristics to substantially refine and filter the data gathered from these social media platforms to compare relative rates within particular contexts that precisely signal use of our target variables. 
These syntactic constructs used to identify the variables rely only on single word precursors (e.g., ``my spouse'') that precisely select the intended uses of the words (as variables) and therefore a bigram model is sufficient for our study.

To focus on interpersonal contexts for the \so variable, we restrict all uses of its variants  to occur only within a possessive pronoun construction, e.g. \textit{my girlfriend}/\textit{his spouse} and later distinguish between first-person and third-person uses, as each carries different social risks. 
For the \pers variable, our focus is on contexts where the gender of the referred persons is inherently ambiguous, i.e., a indefinite referent to a person or group; the underlying hypothesis is that gender need not be ascribed to the referent and any ascribed is a result of underlying attitudes and assumption.
Therefore, we filter uses of the variants to occur immediately after a subset of the determiners, focusing on indefinite articles (e.g. \textit{a dude}), quantifiers (e.g. \textit{most people}), distributives (e.g. \textit{many folks}), and difference words (e.g. \textit{other pals}), as well as broader qualifiers (e.g. \textit{if, when}). 
Indefinite references to ambiguous persons were chosen as opposed to definite ones (e.g. \textit{you guys}) because the latter often takes on a specific audience, which could have a known gender composition that necessitates usage over a ungendered form.

\section{Data}

Our work is drawn from two major social media platforms, Reddit and Twitter, and is focused on English-language conversations in the US. %

\noindent  \textbf{Reddit}
is a major social media platform where individuals participate in communities, known as subreddits, often focused on specific interests, goals, or demographics. Reddit users are primarily English-speakers and recent market research suggests its userbase is largely comprised of American users \cite{statista2020}; we therefore treat content from Reddit as  reflective of this region's attitudes.

On Reddit, communities have formed around particular identities associated with known sociolinguistic variation, e.g., \texttt{r/GayBros}. We treat participation in these communities as an implicit signal of an affiliation with that identity, allowing us to study linguistic variation with respect to these identities. 
Here, we identify four categories of subreddits around identity, with 15 total identities across those categories, shown here each with an example:
\textbf{Politics} Right leaning (r/conservative), Left leaning (r/voteblue); 
\textbf{Religion} General (r/religion), Christianity (r/christianity), Islam (r/islam), Judaism (r/judaism), Non-believers (r/atheism);
\textbf{Sexuality} LGBT+ (r/ainbow), Heterosexual (r/relationships);\footnote{We note that popular subreddits like r/relationships are not exclusively heterosexual; however, the bulk of their content focuses on heterosexual topics and relationships making them an effective contrast to LGBT+-focused communities.}
and \textbf{Gender} Transgender (r/transgender), Men (r/daddit), Women (r/askwomen).\footnote{These  categories were selected to reflect linguistic communities of practice whose style may differ on the basis of lived experience; they do not reflect an exhaustive list of gender identities. Additionally, the inclusion of transgender is not intended as a separate gender and we in no way condone the division of gender identity into these three distinct groups with a disregard of gender fluidity.}

A full list of communities and details of the selection process are provided in supplemental material. %
These categories were chosen based on minimum volume and motivated by prior work showing that individuals modify their language to signal affiliations, attitudes, and beliefs \citep{lakoff2006framing,fausey2010subtle}. Data was collected from all content in these communities during 2013--2019, totalling 73M comments. Additionally, to test for aggregate changes on the platform, we use a random sample of 5M comments per year (35M total) stratified across years. 

\noindent \textbf{Twitter}
is a major international social media platform. Prior work has shown that location data for tweets  can be used to identify lexical variation associated with identity \cite{gonccalves2014crowdsourcing,blodgett2016demographic,abitbol2018socioeconomic}. In these settings, the demographics associated with the location of a tweet  are treated as proxies for the identity of the individual. %
However most tweets do not come with location data; to increase the sample size, we geocode all tweets from a $\sim$10\% random sample spanning 2011-2019 using the method of \citet{compton2014geotagging} and retain only those present in the United States. This method is known to be the least-biased across urban and rural settings \cite{johnson2017effect}, allowing us to study all parts of the US. This geocoding had a median error of 8km in our tests; furthermore, we restricted processing to tweets that were marked as English by Twitter.

To obtain demographic estimates for each tweet's author, we match the inferred location with its containing census tract and use the US Census' 2017 American Community Survey (ACS) variable 5-year estimates. The selected ACS variables focus on socioeconomic status (SES), and cover income, public assistance, education, unemployment, poverty, income inequality, population density, and age dependency. These demographic variables provide a complementary set of indicators for studying variation compared with Reddit.

\myparagraph{Final  Data}
A total of 73M tweets from 2011--2019 and 14M Reddit comments from 2013--2019 are used. Additionally, to test for aggregate changes on the platforms, we use a random sample of 1M  comments per month for Reddit (84M total) and Twitter (108M total). 
The filtering process for the \so variable yields 6.7M contexts for \so from Reddit and 30M from Twitter.
The filtering process for the \pers variable yields 7.3M contexts for \pers from Reddit and 43M from Twitter.
Both filtering processes use NLTK for Part of Speech tagging during preprocessing; 
Full details on terms and filtering are in supplemental material \Sref{variants}.

\section{Measuring Attitudes to Sexuality}

Attitudes about sexuality can be signalled in the use of gender in referring to one's significant other. In settings where heterosexual partnerships are valorized, referring to a significant other of the same sex carries a social and potentially-physical risk from admitting to being different from heterosexual practices \cite{wilkinson2015sexuality,cadieux2015you}. To minimize this risk, some LGBT+ individuals refer to significant others using the variant form \textit{partner}, which leaves the gender status ambiguous.
However, use of partner \textit{only} by only LGBT+ community members would result in it being a clear marker of a marginalized community. Therefore, allies of this community are known to use this word to decrease its association with a marginalized identity \cite{kitchener_2019}.
Thus, uses of gender markers to refer to significant others reflect underlying attitudes of acceptance towards gay marriage, allowing us to study changes in these attitudes by examining relative rates among variants' uses.

Here, we ask to what degree is this shift in attitude mirrored in changes in language use, testing two hypotheses. \textbf{H1}: LGBT+ communities will increasingly use gendered terms when referring to relationship status. 
\textbf{H2}: Heterosexual individuals will increasingly use gender-neutral markers in \so. 

H1 is motivated by \citet{heisterkamp2016challenging}, who in a small observational study found that LGBT+ individuals preferred using gendered markers, which suggests that use of gender-neutral markers by the LGBT+ members may actually be in the minority.
To test these hypotheses, we measure variant use longitudinally and across identities.
To avoid larger communities outweighing smaller ones and contributing more to an observed change in progress, in all cross-community comparisons, we controlled for community size by bootstrapping the mean probability within each category of subreddits, and show the 95\% confidence intervals in the figures.

\begin{figure}[!t]
    \centering
    \raisebox{-0.5\height}{\includegraphics[width=0.23\textwidth]{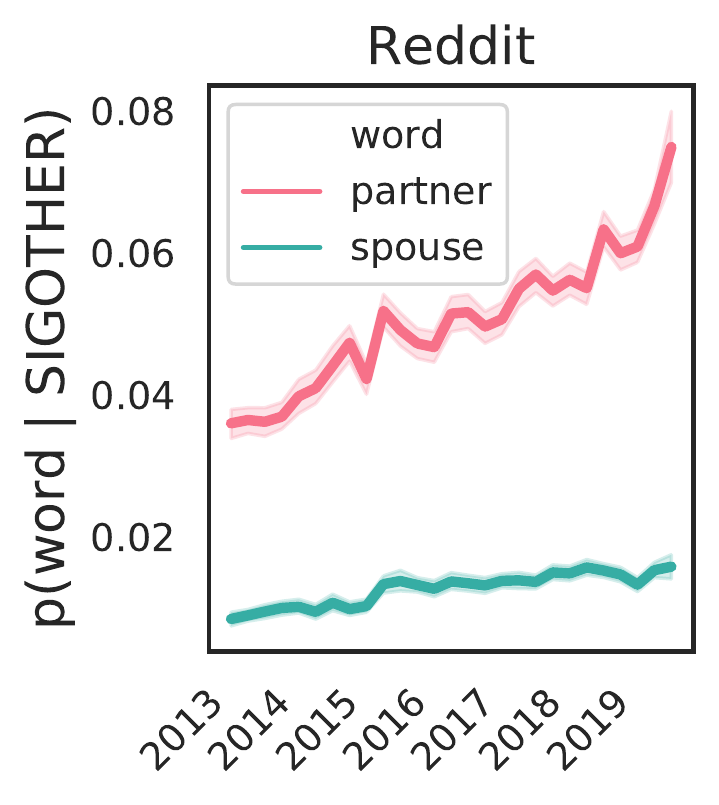}}
    \raisebox{-0.5\height}{\includegraphics[width=0.23\textwidth]{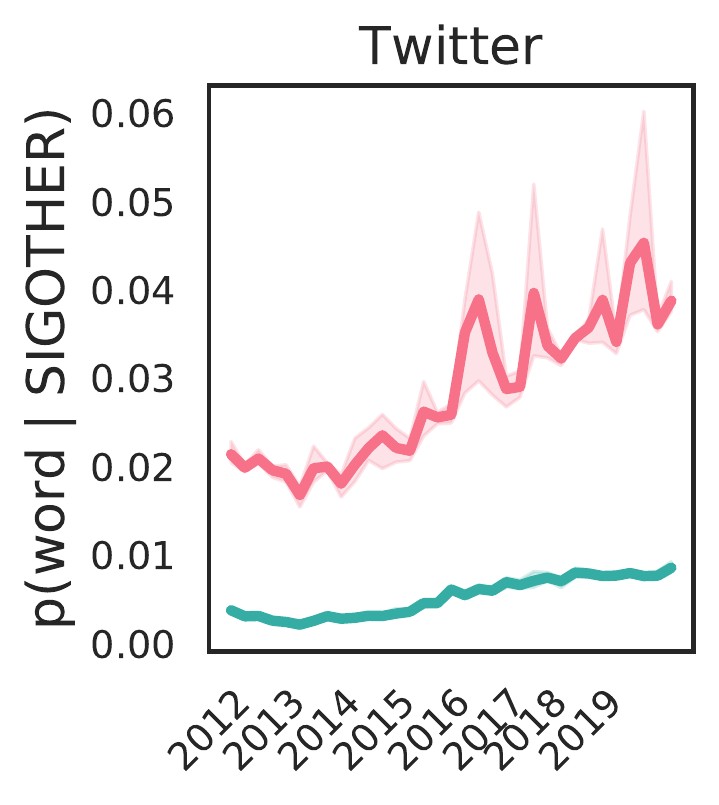}} 
    \caption{Probabilities of using \so \textit{partner} and \textit{spouse} terms across all of Reddit and Twitter, over time; both terms increase in use across platforms.} 
    \label{fig:so-reddit-twitter}
\end{figure}

\begin{figure*}[!t]
    \centering
    \begin{tabular}{cccc}
    \subfloat[Sexuality]{\includegraphics[width=0.22\textwidth]{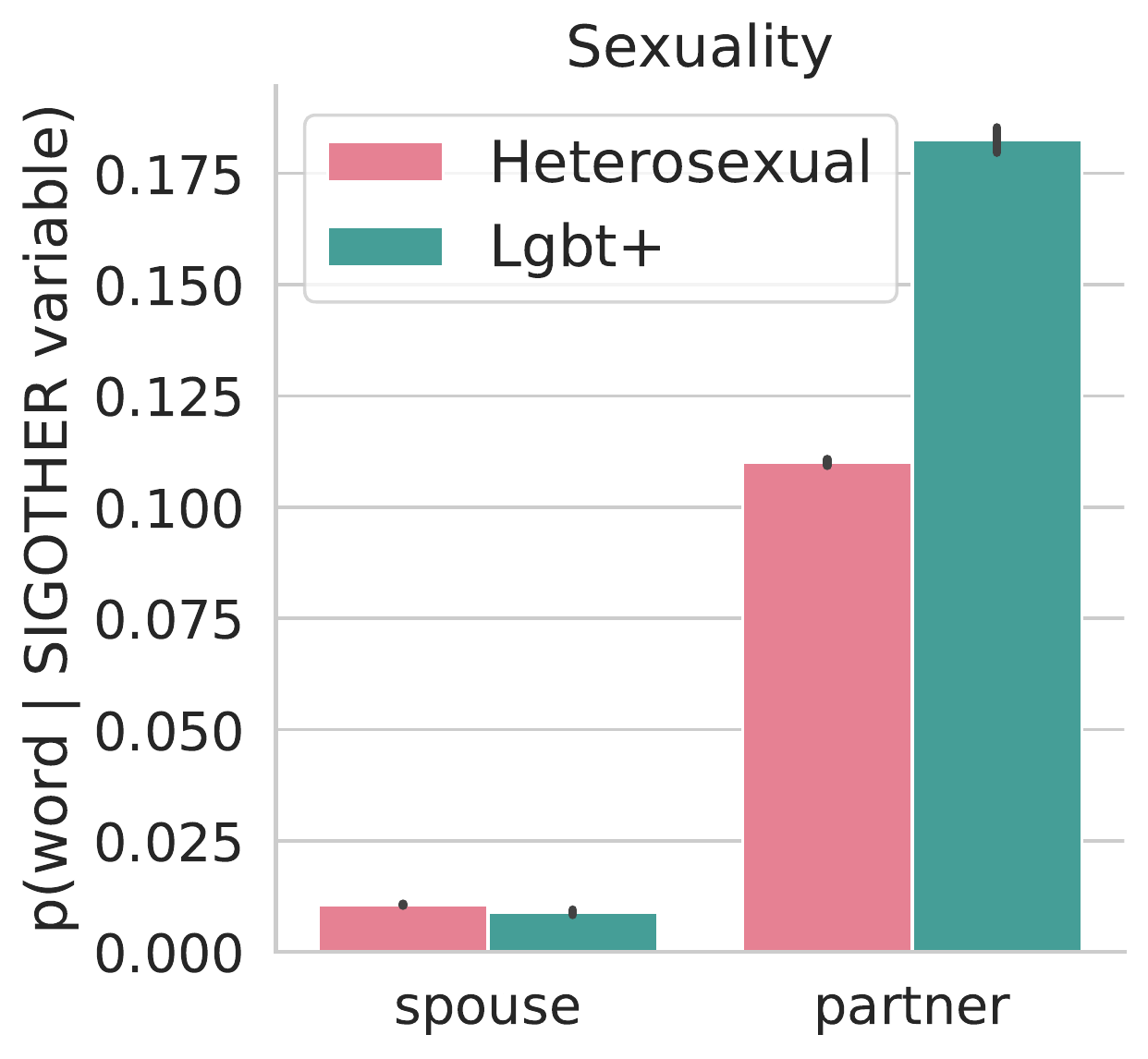}} & 
    \subfloat[Gender]{\includegraphics[width=0.22\textwidth]{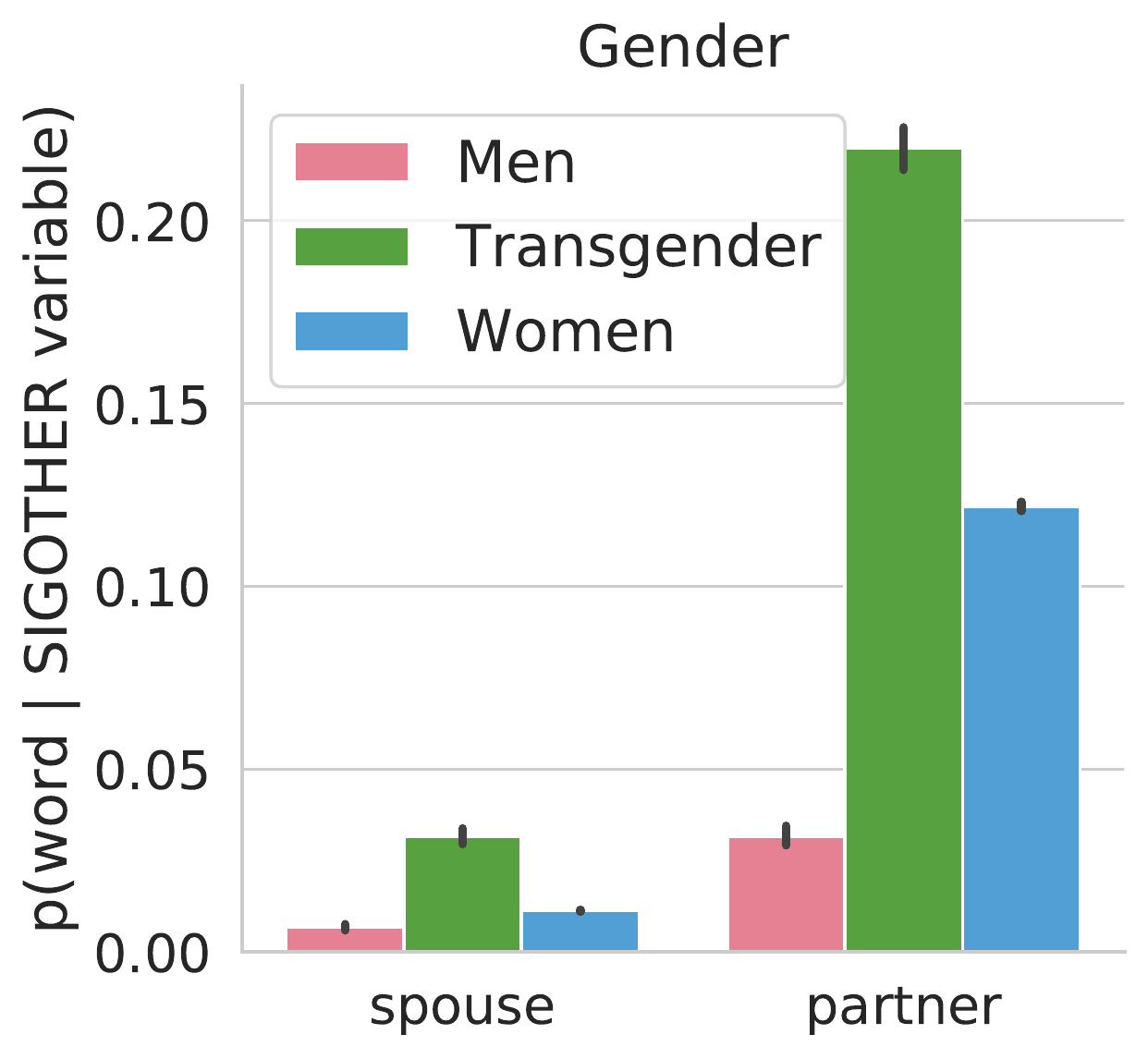}} &  
    \subfloat[Political Leaning]{\includegraphics[width=0.22\textwidth]{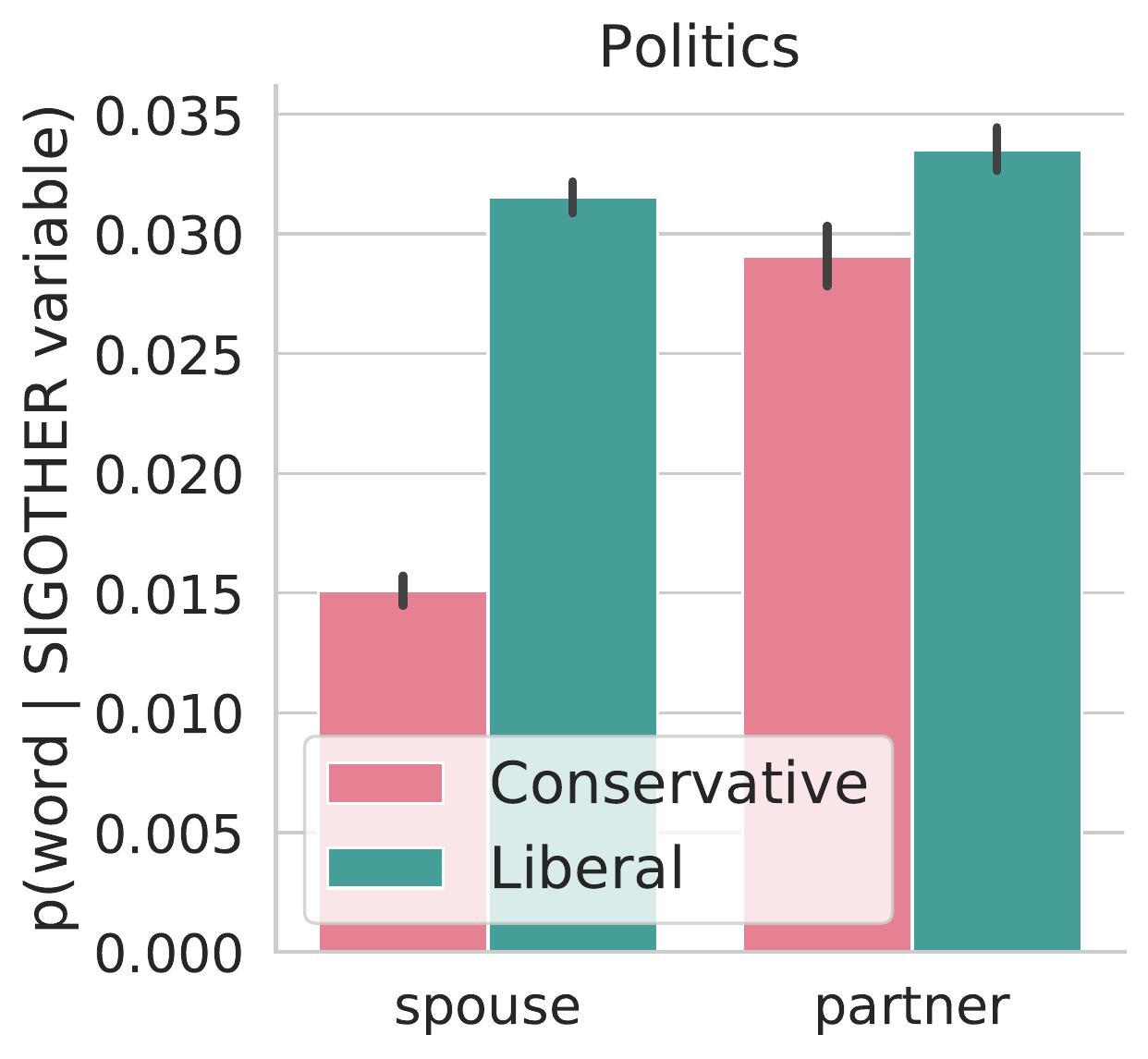}} &    
    \subfloat[Education]{\includegraphics[width=0.22\textwidth]{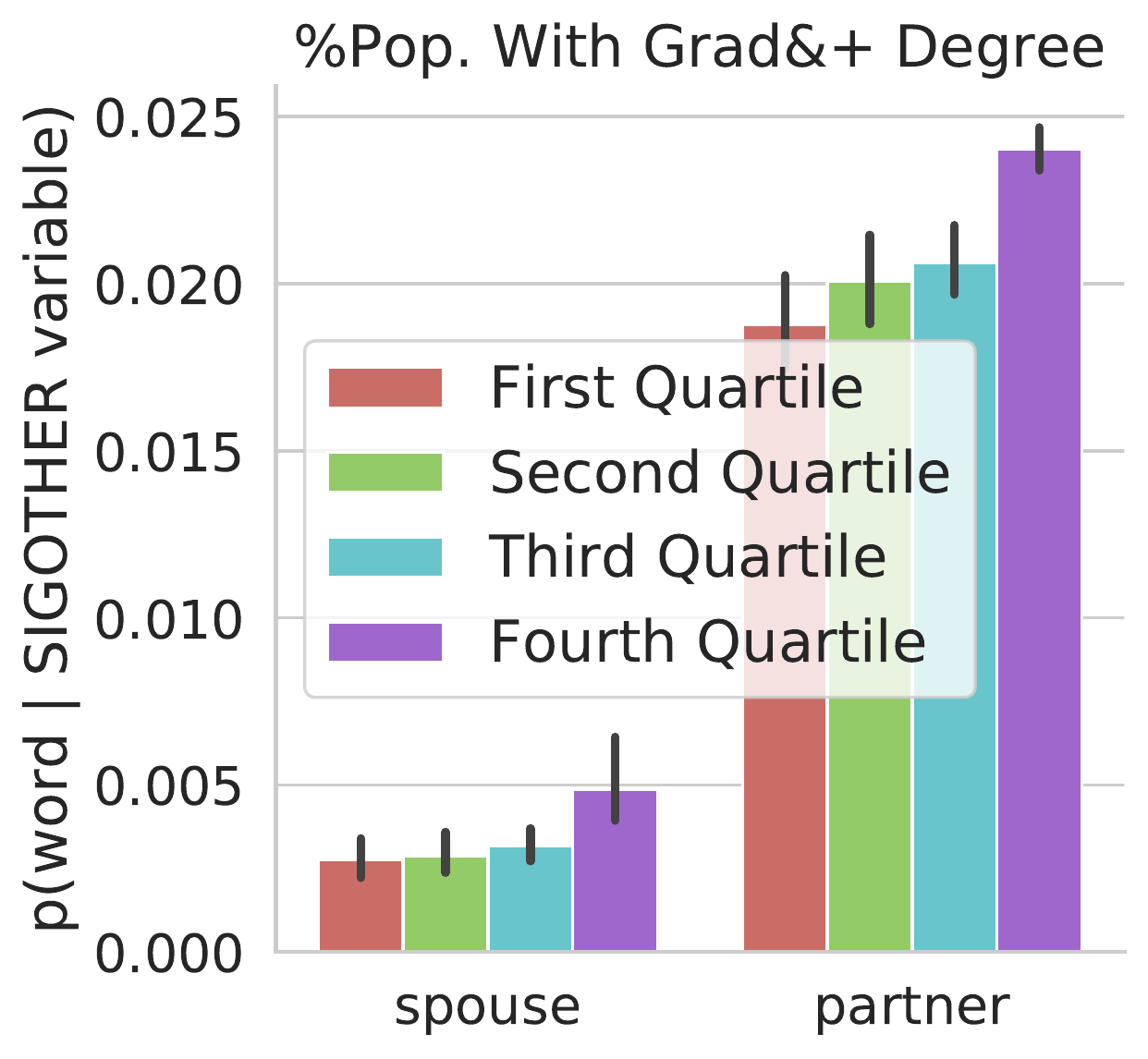}} \\
    \subfloat[Sexuality]{\includegraphics[width=0.24\textwidth]{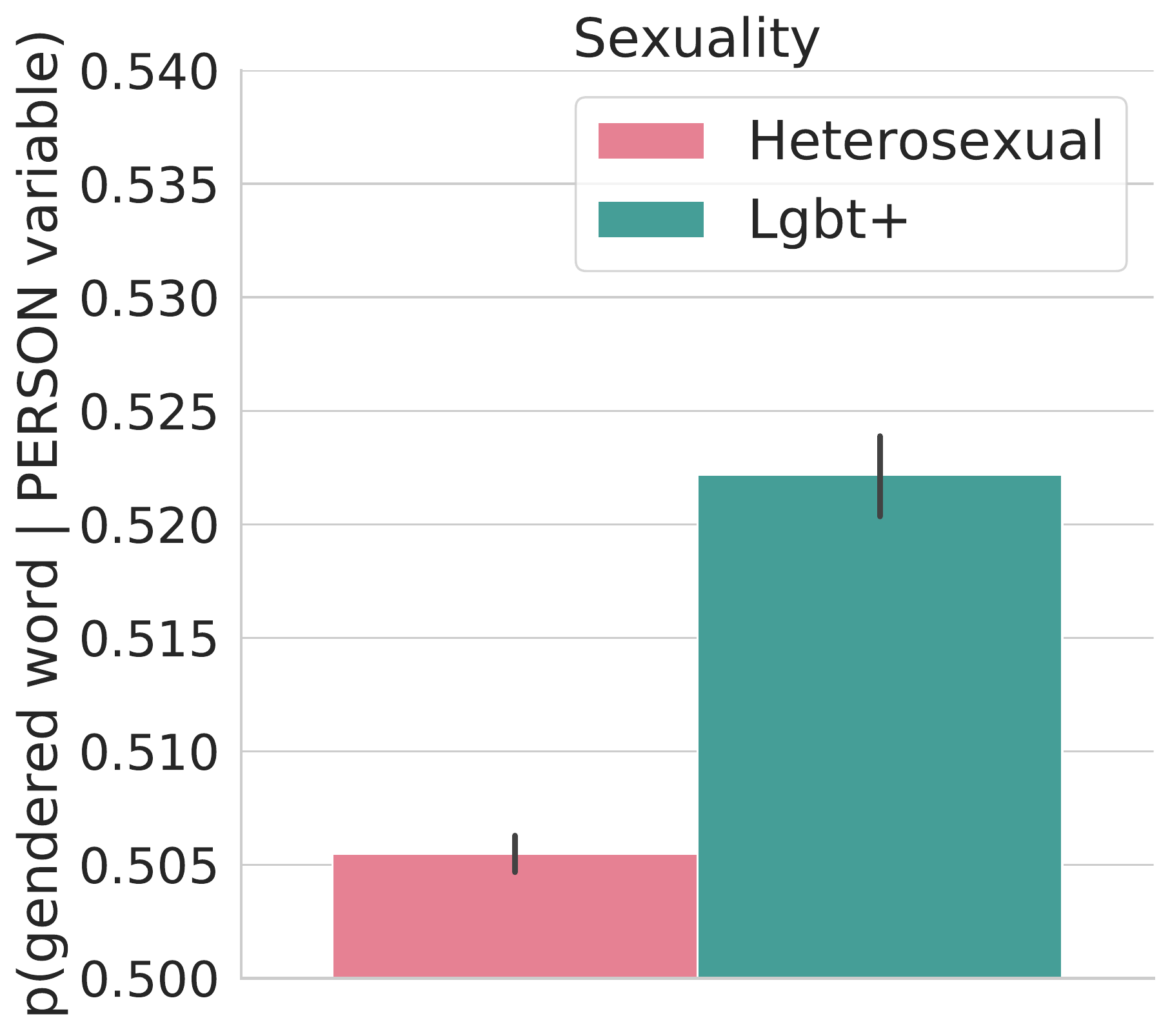}} & 
    \subfloat[Gender]{\includegraphics[width=0.24\textwidth]{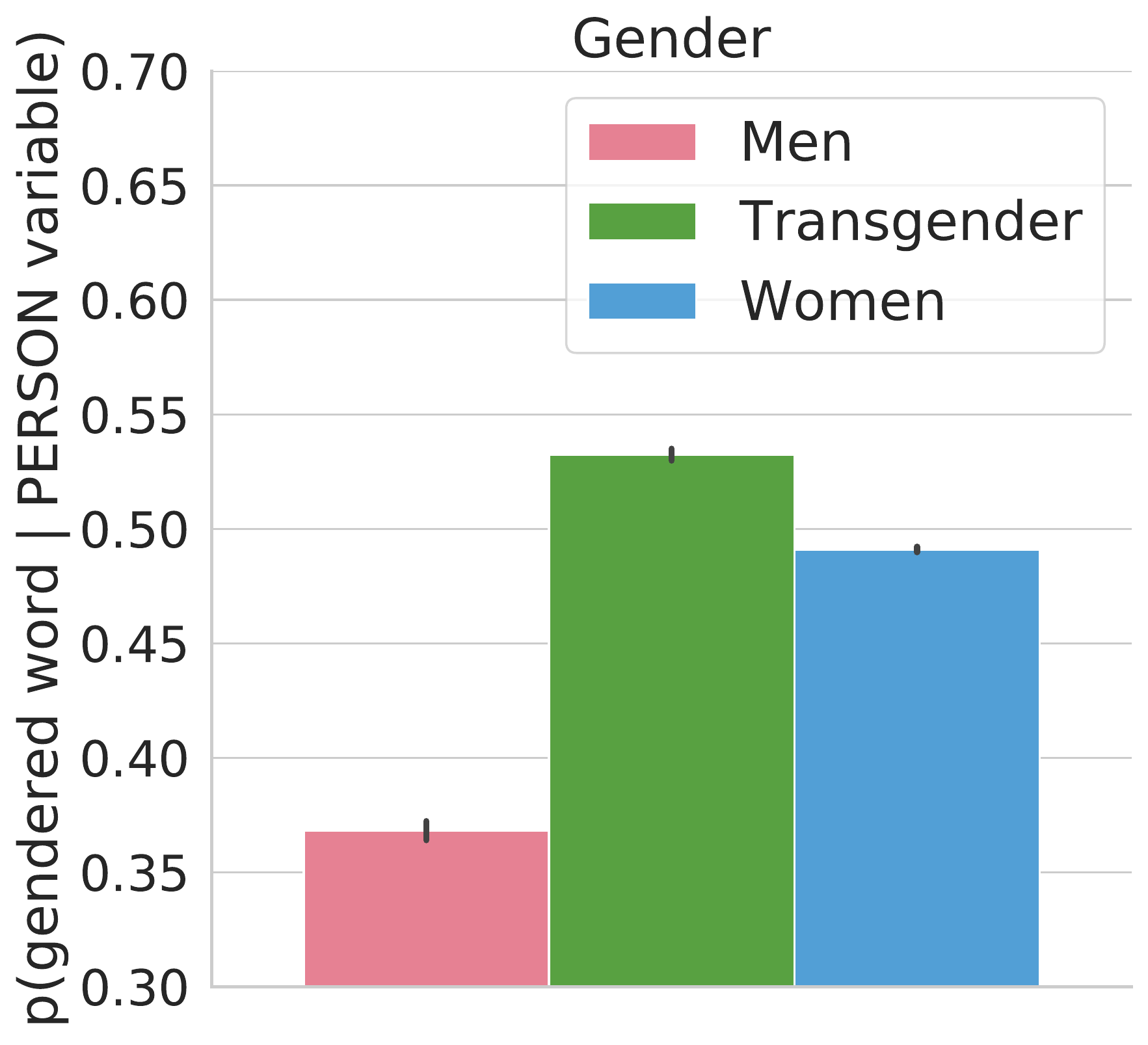}} &
    \subfloat[Political Leaning]{\includegraphics[width=0.24\textwidth]{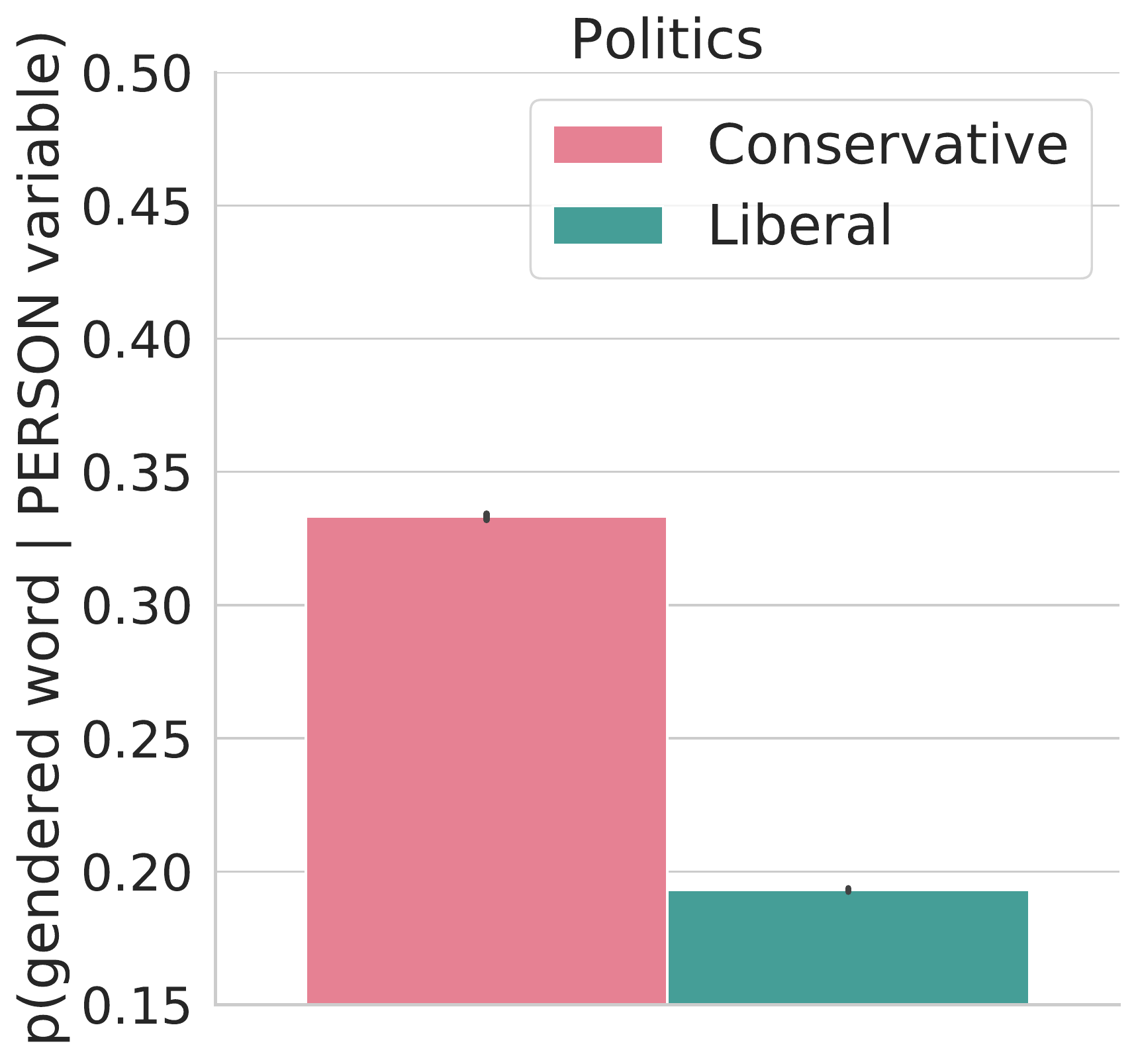}} &    
    \subfloat[Income]{\includegraphics[width=0.24\textwidth]{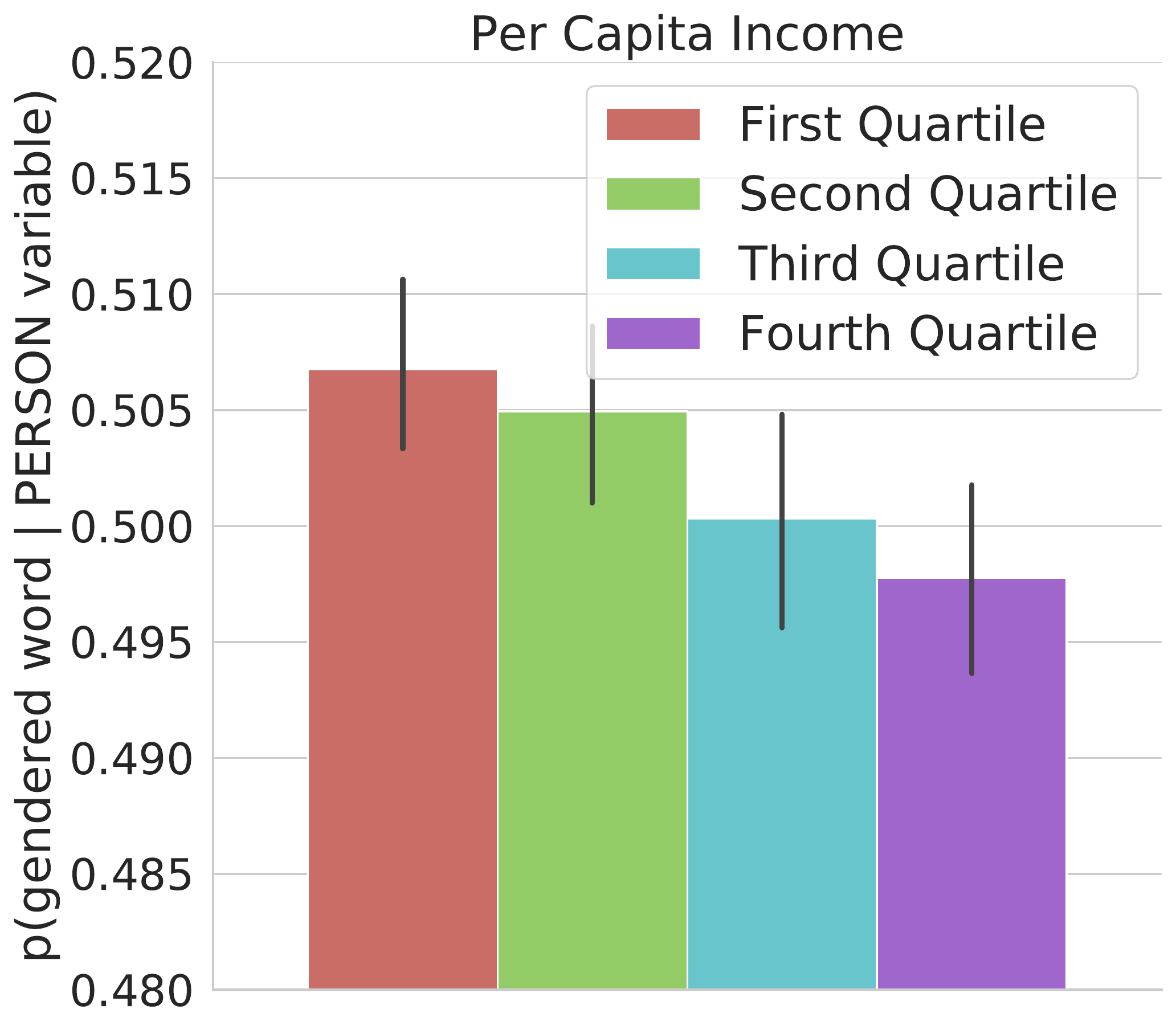}}

    \end{tabular}    

    \caption{ (a)-(d) The probabilities of \textit{partner} and \textit{spouse} use in \so within identity-aligned communities shows greater use in non-heterosexual sexual orientation \& gender communities than in their hetero-normative \& traditional opposites, as well as in regions with higher education and  socioeconomic status. 95\% confidence intervals are shown. (e)-(h) Probabilities of using \textbf{gendered} \pers terms across Reddit identity-affiliated communities; additional figures for Twitter demographics are shown in supplemental material. 
    }
    \label{fig:who-uses-so}\label{fig:who-uses-pers}
\end{figure*}

\subsection{Changing Use in \so}
\label{sec:so}

Lexical change mirrors the corresponding increasing acceptance of same-sex marriage in the US \cite{ofosu2019same,twenge2020increased}, with both Reddit (\textit{partner}: $r$=0.950, $p$<0.01; \textit{spouse}: $r$=0.916, $p$<0.01) and Twitter (\textit{partner}: $r$=0.901, $p$<0.01; \textit{spouse}: $r$=0.943, $p$<0.01) having increased rates of gender-neutral markers of \so (\fref{fig:so-reddit-twitter}). While gendered variants still account for the majority of uses, this trend signals an underlying change of attitude by reducing the focus on gender in describing a \so. To understand the mechanisms behind this change, we examine who is likely to use these alternates and any changes in behavior.

\myparagraph{Who uses non-gendered variants?}
To measure association with identity, on Reddit, we measure the relative rates within each group of subreddits associated with an identity, and on Twitter, we stratify census tracts for pertinent identities (e.g., education) into quartiles to show relative difference between high and low-valued areas. Measurements are taken over all of the data and, here, we show the rates for \textit{partner} and \textit{spouse}, which are the two most common non-gendered variants.

The result, shown in \fref{fig:who-uses-so}, reveals four trends. First, confirming prior expectations around the association of these words with sexuality, the gender neutral forms are used most frequently by non-heterosexual communities \cite{heisterkamp2016challenging}. 
Second, communities for gender identities other than male and female have substantially larger use of gender-neutral forms; this language likely reflects a concerted effort towards inclusivity within a marginalized community that also has partial overlap with LGBT+ communities. %
Liberal communities, which in the US are known to have more favorable attitudes towards gay marriage \cite{sherkat2011religion}, exceed that of uses in conservative communities. %
Higher SES, as shown through income, education, and public assistance, see greater adoption than their lower SES counterparts, mirrored in recent findings on SES and their favorability towards gay marriage \cite{jakobsson2013attitudes,anderson201421st}. Results for variable usage by density, income and inequality were found to be similar; for brevity, education is shown in Figure \ref{fig:who-uses-so}. Complementary figures are in Supplemental Material section \ref{expanded-results}.
The urban-rural divide reflects both (i) known attitudes of rural communities that typically placed a heightened value on ``traditional moral standards'' \cite{bell1995queer} which would disfavor the language of LGBT+ communities and (ii) a self-selection of LGBT+ people to denser, urban areas \cite{gorman2014mobile}. %

Despite known prejudices towards LGBT+ identities by some religious denominations \cite{besen2007young,fetner2008religious}, uses of partner and spouse were higher in categories of communities containing these denominations than in non-believing communities (e.g., r/atheism), mirroring studies showing that progressive movements within non-believing communities like Atheism+ are still in the minority \cite{kettell2014divided}.

\myparagraph{How has gender-signalling changed?}
\begin{figure*}[!t]
    \centering
    \begin{tabular}{ccc}
    \subfloat[(1P) Gendered  w.r.t.~Sexuality ]{\includegraphics[width=0.29\textwidth]{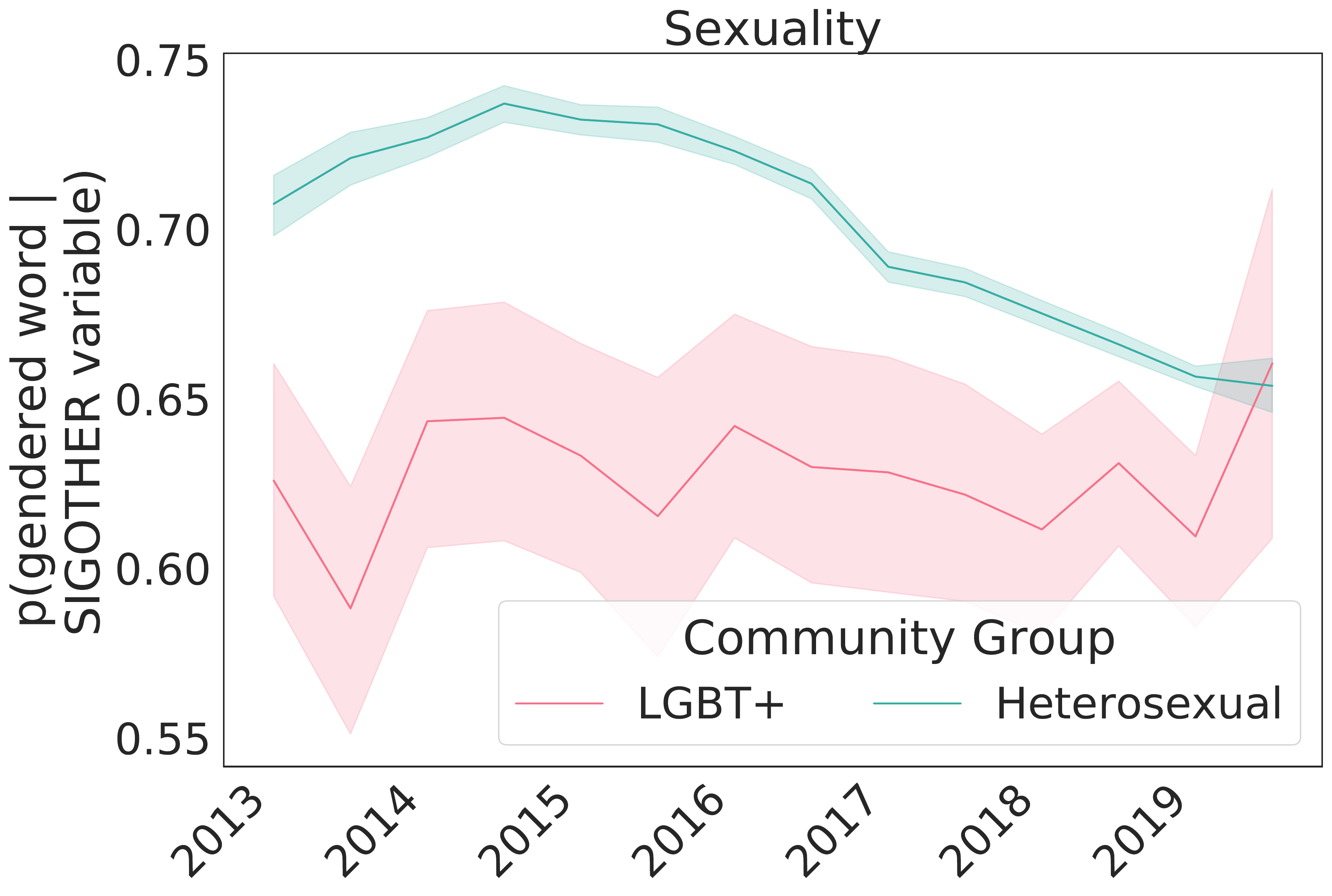}} & 
    \subfloat[(3P) Gender-Neutral  w.r.t.~Sexuality ]{\includegraphics[width=0.29\textwidth]{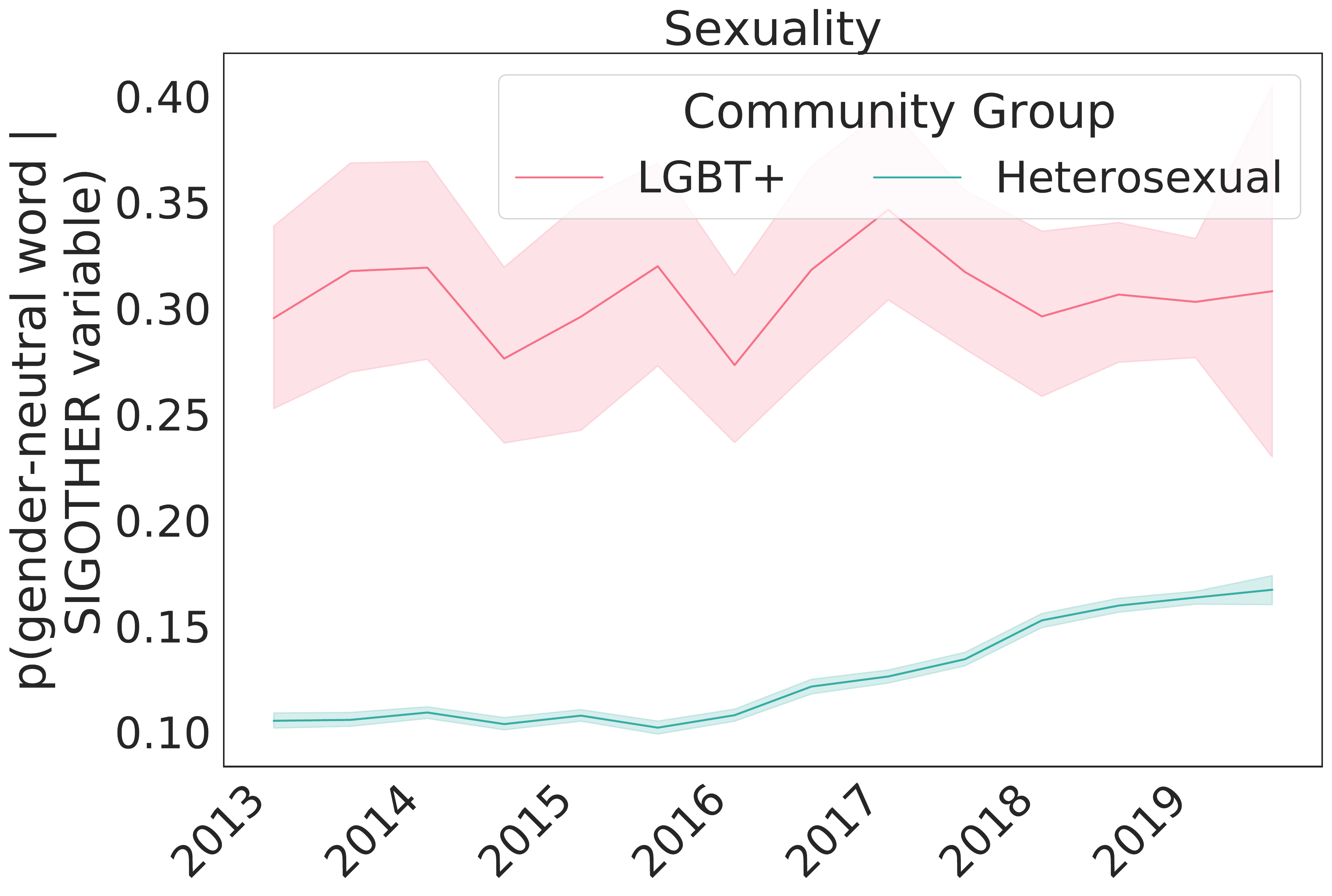}} &
    \subfloat[(1P) Gender-Neutral  w.r.t.~Gender ]{\includegraphics[width=0.29\textwidth]{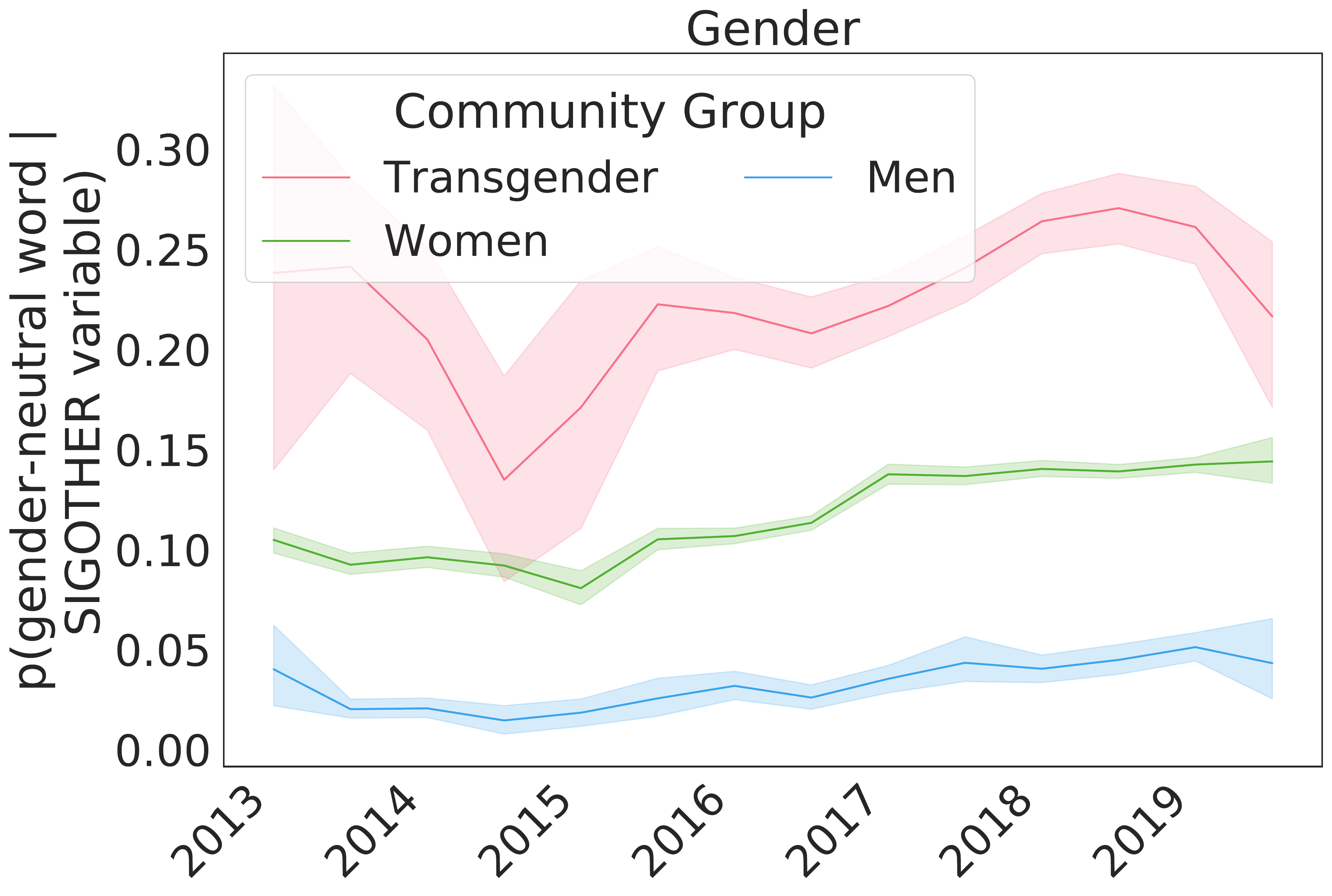}} \\
    
    \end{tabular}    
    
    \caption{ Changes in the probabilities of using \so terms with respect to (w.r.t.) sexuality and gender communities show heterosexual and traditional gender identity communities adopting the gender-neutral term of partner at more rapid rates than their nontraditional counterparts. Plots show either first-person (1P) use or third-person (3P) use of \so. } 
    \label{fig:so-temporal}
\end{figure*}
Our results demonstrate strong association of gendered and non-gendered terms with identity, in line with observational studies. We now test our two hypotheses by examining changes in \so variant usage over time with respect to these identities by aggregating the most common six gendered  (girlfriend, boyfriend, wife, husband) and ungendered (partner, spouse) markers on Reddit, controlling with respect to first-person (e.g. \textit{my wife}) and third-person (e.g. \textit{her boyfriend}) use. We explicitly note this distinction as first-person uses of the \so term may carry different social penalties for different individuals in various communities. For example, a gay man referring to his partner as ``my boyfriend'' is a risky act of self-disclosure, while a person discussing another person's relationship and referencing ``his boyfriend'' (rather than ``his partner'') is taking less of a personal risk, instead showing acceptance of gender-neutral referents in the individual's common lexicon.

\fref{fig:so-temporal} shows changing rates for the three settings related to sexuality and gender needed to test the hypotheses; plots for all others identities are shown in supplemental material.
Figures \ref{fig:so-temporal} (a) and (b) support \textbf{H1}: Across the LGBT+ communities, the rate of gendered markers continues to exceed that of ungendered markers across time, and shows no statistically significant trend of change (1P: $r$=0.156, $p$=0.594; 3P: $r$=0.126, $p$=0.667). Through studying person-reference practices across large-scale social communities, we validate \citet{heisterkamp2016challenging}'s concentrated findings that usage of such referents in LGBT+-community contexts continues to show a resistance to, and a divergence from, heteronormative social constructs. 
Figures \ref{fig:so-temporal}(a) and (b) support \textbf{H2}, with substantial increases in non-gendered terms among heterosexual communities for both third ($r$=0.917, $p$<0.01) and first-person ($r$=0.982, $p$<0.01) use. While the Reddit communities used to identify this trend (e.g., r/relationships) do contain some LGBT+ members, the lack of substantial increases in non-gendered forms within LGBT+ subreddits suggests that the shift is due to changing attitudes within the heterosexual community.

As a follow up study, we test whether these changes are driven by a particular gender. Pearson's correlation is computed alongside statistical significance testing over bootstrapped mean probabilities calculated across 3-month intervals. Figure \ref{fig:so-temporal} (c) shows that all gender communities in our study increased their rates of non-gendered markers with women-focused ($r$=0.884, $p$<0.01) and men-focused ($r$=0.731, $p$<0.01) communities increasing more than transgender ones ($r$=0.444, $p$=0.11), suggesting wide-spread normalization.

We argue that the increased use of a LGBT+-marked term ``partner'' by non-LGBT+ community members is an example of dialect merging where the dominant identity (here, heterosexual) adopts the language of the minority as a standard. This trend draws parallels with the adoption of African American Vernacular English (AAVE) by white Americans \cite{cutler1999yorkville}. However, race-based linguistic markers face problematic adoption due to perceptions of who is a member of the community and appropriately use its language \cite{sweetland2002unexpected}. In contrast, the gay community includes allies, which potentially license this linguistic adoption; though we note that, as a marker of in-community status, this adoption by individuals outside the community has met some resistance \citep{romack_2018,werder_lee_sulc_howse_2017}, raising the question of whether this behavior is linguistic adoption versus appropriation.
In addition, our results point to an absence of lexical leveling \cite{milroy2002introduction,kerswill2003dialect}, where the language of a minority community is gradually replaced by that of the majority as the minority is integrated; often, individuals in a minority linguistic group assimilate to the mainstream usage (leveling) due to the associated perceived prestige, but here this trend is reversed.

\subsection{Effects of Marriage Equality Acts}

During the time period of 2004 to 2015, multiple states in the United States passed Marriage Equality Acts (MEAs) that allowed LGBT+ couples to legally marry. As a result, marriage rates for these couples rose substantially and  passage of the acts was shown to increase social acceptance \cite{ofosu2019same}. These passages provide an ideal setting for a natural experiment to test whether legalization influenced linguistic choice.

\myparagraph{Methods}
To analyze the effect of passage of MEAs, we construct a  difference-in-differences (diff-in-diff) model as a quasi-causal analysis of the effect on linguistic choice.
In a small-scale interview-based study, \citet{digregorio2019same} found that passage of an MEA did not mean the traditional language of marriage would be adopted, suggesting we should observe no change in certain \so forms. Therefore, we test specifically for changes in spousal terms---partner, spouse, wife, and husband---on whether individuals who marry after the passage will use the gendered or gender neutral forms.
Since states pass MEAs at different times, we adopt a staggered diff-in-diff formulation that controls for changes in usage across real time, while measuring the changes relative to treatment \cite[cf.][]{stevenson2006bargaining,gipper2020public}. 
This model is formalized as
\[
y_{ij} = \alpha_i + \lambda_j + \sum_{j=m}^{-1}\pi_j T_{ij} + \sum_{j=1}^g\phi_j K_{ij} + \epsilon_{ij}
\]
where $y_{ij}$ is the probability of using a particular form of the \so variable, $\alpha$ and $\lambda$ are variables for the state and absolute time (as month) of measurement, which account for baseline changes in the rates of words' uses over time and across states. $K$ and $T$ are pre- and post-treatment interactions of the relative month offset from a state passing an MEA and a dummy variable indicating passage of an MEA; the fitted $\pi_j$ and $\phi_j$ parameters reveal the effect of treatment on the outcome variable, i.e., the particular \so word used.  We use a twelve month period around the passage, setting $m$=-12 and $g$=12, to assess trends.

Data for the diff-in-diff model is selected from all tweets referring to the \so variable in the twelve months before and after the passage of the MEA in a state.
Tweets were then filtered to the 30 states passing a MEA within our dataset's timespan; a total of 6.7 million tweets are included in this analysis.

\begin{figure}[t]
  \centering
  \includegraphics[width=0.23\textwidth]{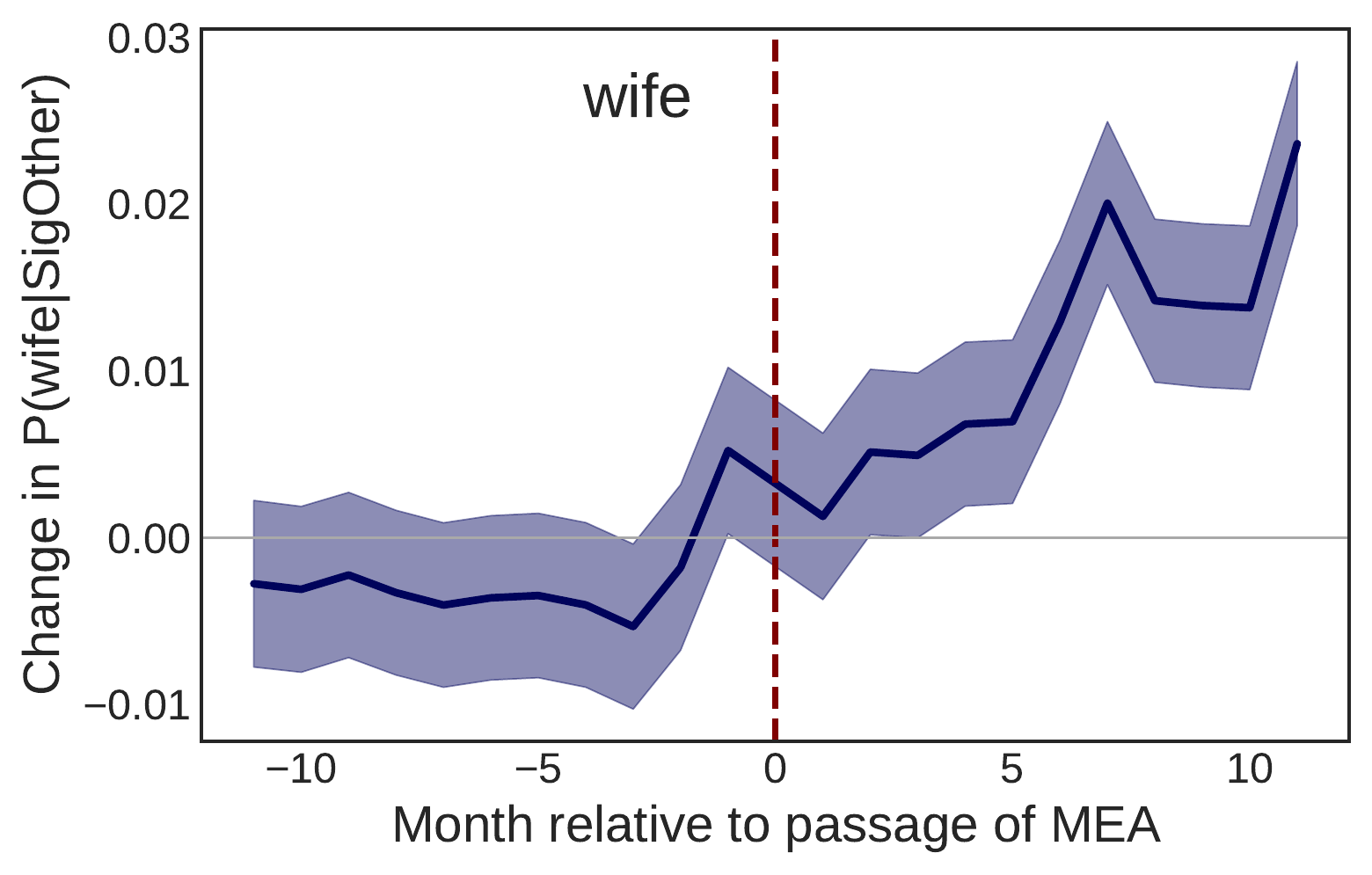}
  \includegraphics[width=0.23\textwidth]{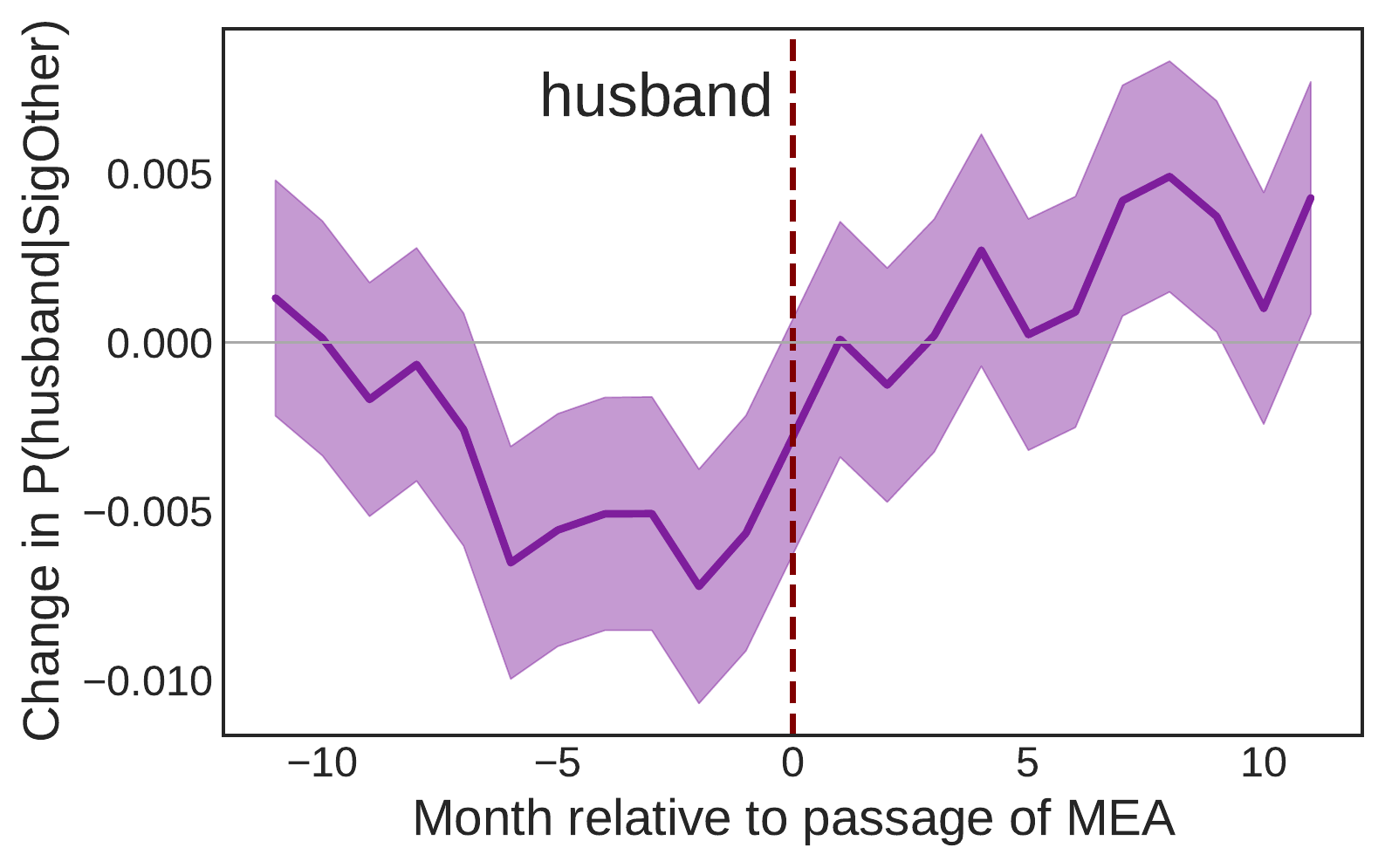} \\
  \includegraphics[width=0.23\textwidth]{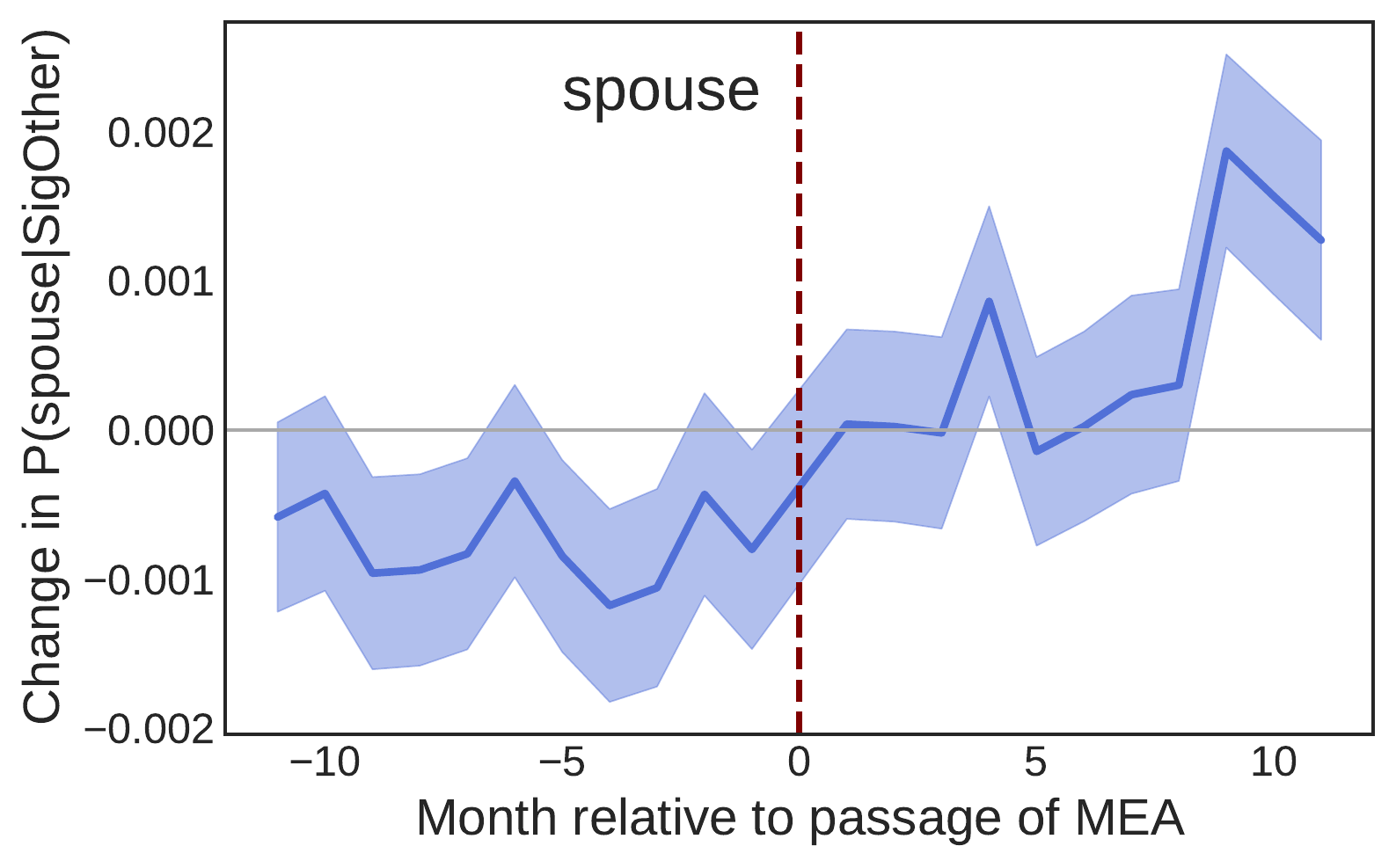}
  \includegraphics[width=0.23\textwidth]{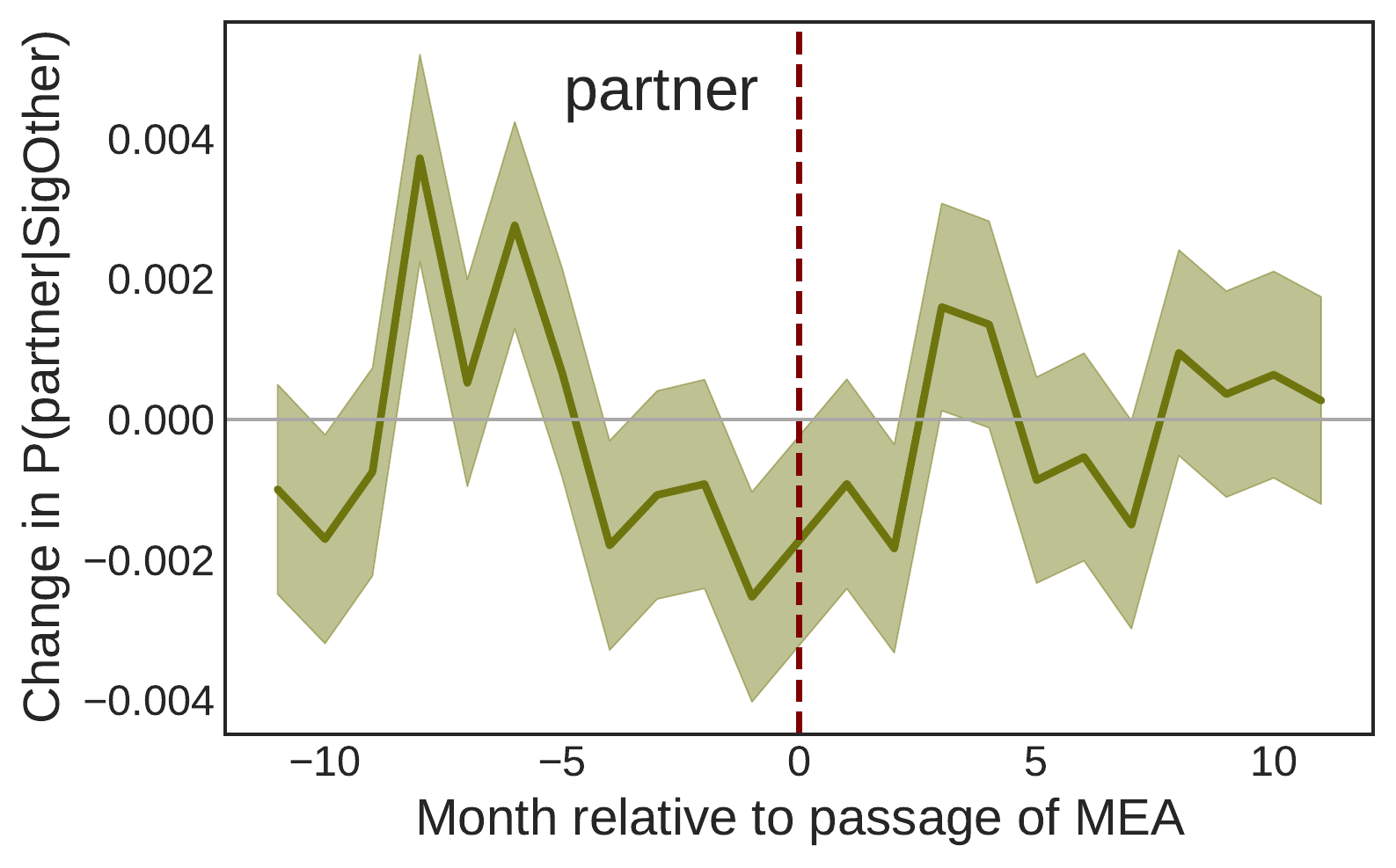}
  \caption{ Estimate effects of passage of an MEA on uses of spousal terms in the \so variable shows that passage resulted in a sharp increase for traditional gendered terms, but no increase in the use of gender-neutral partner. Shaded regions show standard errors. }
  \label{fig:mea-effect}
\end{figure}

\myparagraph{Results}
As shown in Figure \ref{fig:mea-effect}, the rates of marital terms in the \so variable substantially increased after the passage of a MEA, with the largest absolute increases in gendered markers, particularly for \textit{wife}. 
Note that the diff-in-diff model controls for baseline changes in usage by month and state, which mitigates potential confounds from overall fluctuations in how these terms are used.
Our findings run counter to the study of \citet{digregorio2019same}, where by computationally examining larger data across multiple states, we find the opposite result: after passage, LGBT+ couples are much more likely to use gendered spousal terms or use the traditional non-gendered term \textit{spouse} rather than partner. While the passage of an MEA likely facilitates the use of marriage-related gendered terms, the underlying causes in this linguistic change are likely much more complex and due to changing attitudes and the efforts of LGBT social movements in securing the passage of the MEA itself.
Our findings suggest a decreased social penalty for explicitly stating one's sexuality (via a gendered \so term) from increased  acceptance.

\section{Attitudes on Gender Equality}

Traditionally, Standard American English has been gendered in its referents to people, with phrases like ``guys'' referring to both male and mix-gendered groups \cite{mclennan2004guy}. Studies have argued that these marking practices reflect latent bias in gender expectations and reinforce masculinity as the normative gender \cite{wilke1994women,connell2005hegemonic}. Recent efforts have pushed for increased use of gender-neutral forms \cite{schweikart1998gender}, where variants like ``people'' or ``folks'' are used instead.\footnote{These reference terms are in addition to complementary adoption of gender-neutral pronouns such as ``they'' in English \citep{bodine1975androcentrism,lascotte2016singular} or ``hen'' in Swedish \cite{gustafsson2015introducing}, which are outside the scope of the \pers variable but whose adoption is likely also reflective of changing attitudes.}
Using our data, we test for whether these efforts have had an effect and which groups are driving linguistic change. %

\subsection{Changing Uses in Gendered Markers}

\begin{figure}[!t]
    \centering
    
    \raisebox{-0.5\height}{\includegraphics[width=0.23\textwidth]{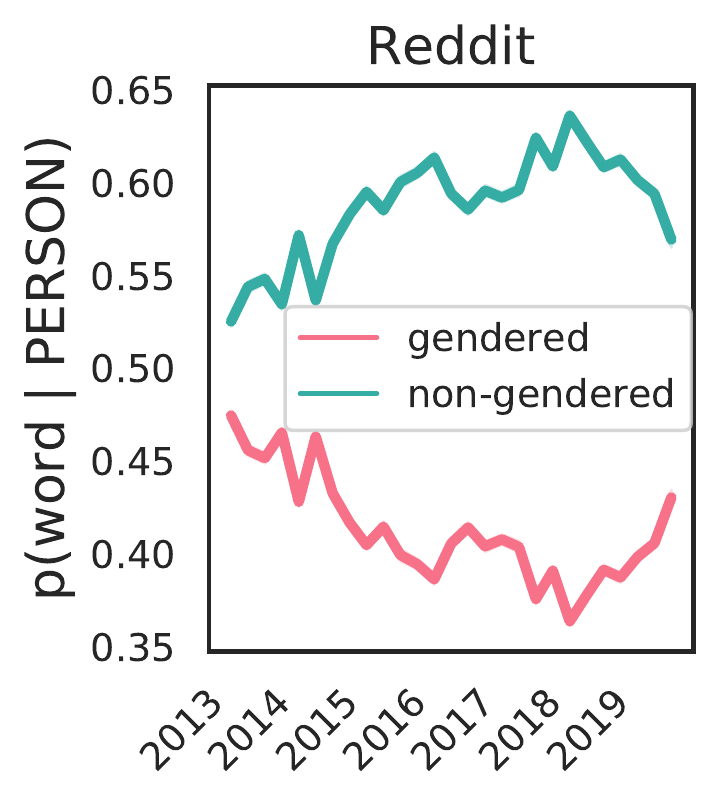}}
    \raisebox{-0.5\height}{\includegraphics[width=0.23\textwidth]{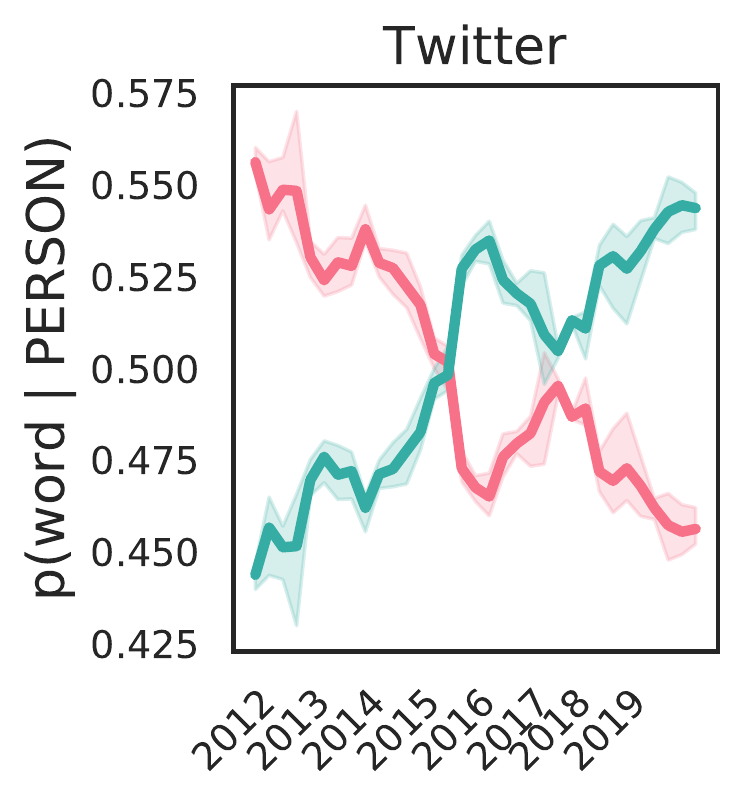}}
    \caption{Probabilities of using gendered and non-gendered \pers terms across all of Reddit and Twitter, over time; non-gendered terms see an increase in use across both platforms.}
    \label{fig:pers-reddit-twitter}
\end{figure}

\myparagraph{Is there change?}
As an initial test, we plot the relative rates of gendered and non-gendered variants of \pers from random samples on both platforms, restricting Twitter to US locations. Shown in \fref{fig:pers-reddit-twitter}, individuals in these settings increasingly use gender-neutral terms to refer to people. Both platforms show consistent trends suggesting that American English is indeed becoming more gender neutral---Pearson $r$ for non-gendered referents are $r$=0.771, $p$<0.01 and $r$=0.966, $p$<0.01 for Reddit and Twitter, respectively.

\myparagraph{Who uses gendered markers?}
To test for broad association with gendered uses of \pers, we compute the relative rates of gendered and non-gendered markers for each identity group on Reddit and use the ACS to estimate demographics on Twitter. %
The results, shown in \fref{fig:who-uses-pers}, reveal three notable trends.
Minority communities around sexual and gender identities are more likely to use gendered language, with the exception of asexual communities (Supplemental \Sref{asexual-case-study}).  

Results on political affiliation show the strongest differences. Liberal communities are less likely to use gendered language than their conservative counterparts, mirroring norms around gendered roles and expectations associated with each party \cite{lakoff2010moral}. 

Following studies on attempts in educational and workplace settings \cite{olgiati2002promoting,pauwels2005education} to actively promote gender equality and the use of gender-neutral language, 
higher income quartiles show a statistically-significant change towards use of gender-neutral language ($p<0.01$ via Kolmogorov-Smirnov) difference in term usage than lower quartiles. Complementary results following expectations on urban density, education, and inequality SES indicators are shown in Supplemental Material section \ref{expanded-results}. All showed statistically-significant differences among the quartiles, except for education.

\myparagraph{Who drives this change?}
Among all categories, the sharpest overarching decrease in gendered marker use is seen in the gender identities studied, for men ($r$=-0.14, $p$<0.01), women ($r$=-0.38, $p$<0.01), and transgender ($r$=-0.29, $p$<0.01) communities, using the same correlation and significant testing calculations in \sref{sec:so}.

\begin{figure}[ht]
    \centering
    \includegraphics[width=0.48\textwidth]{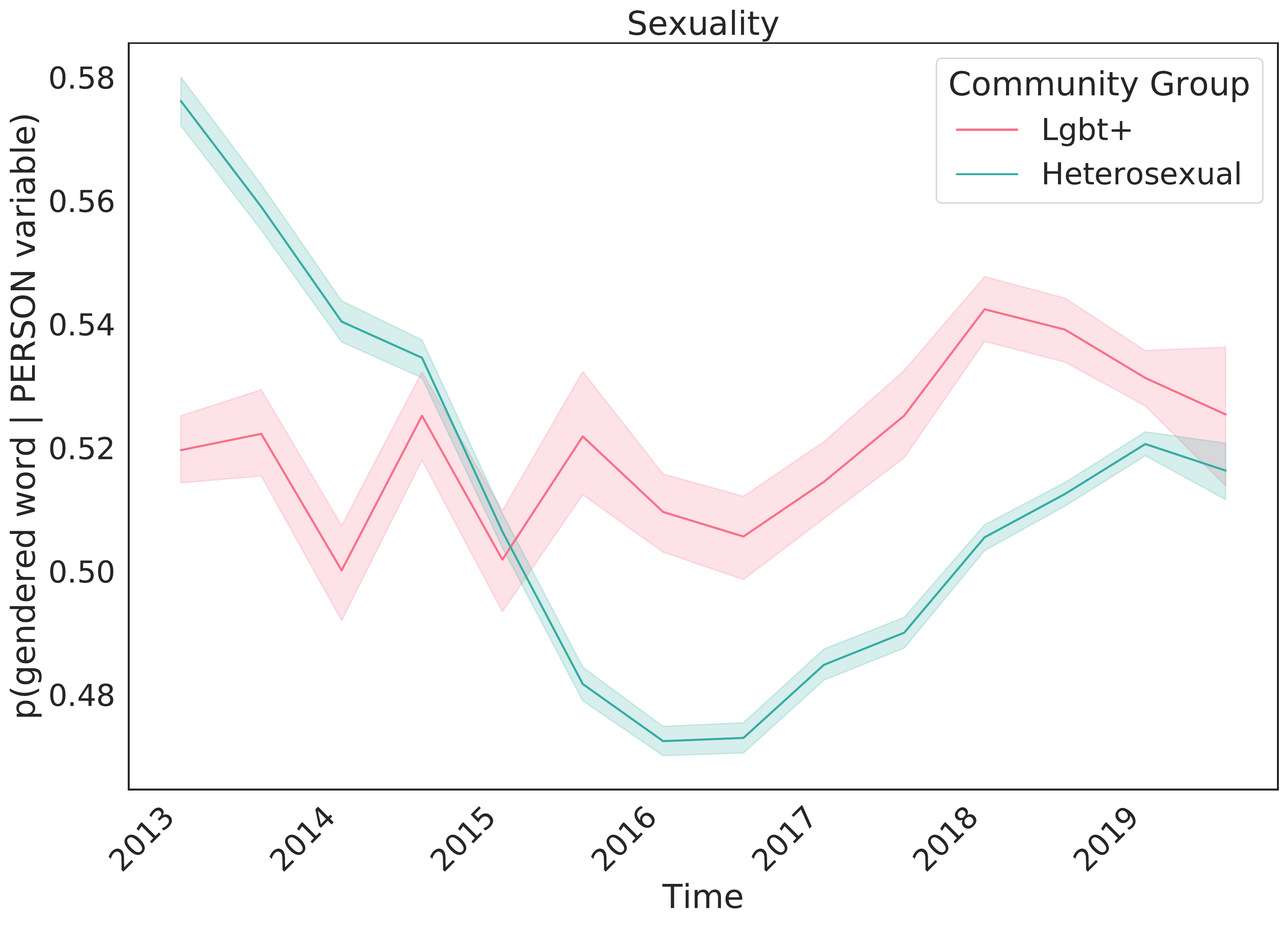}
    \includegraphics[width=0.48\textwidth]{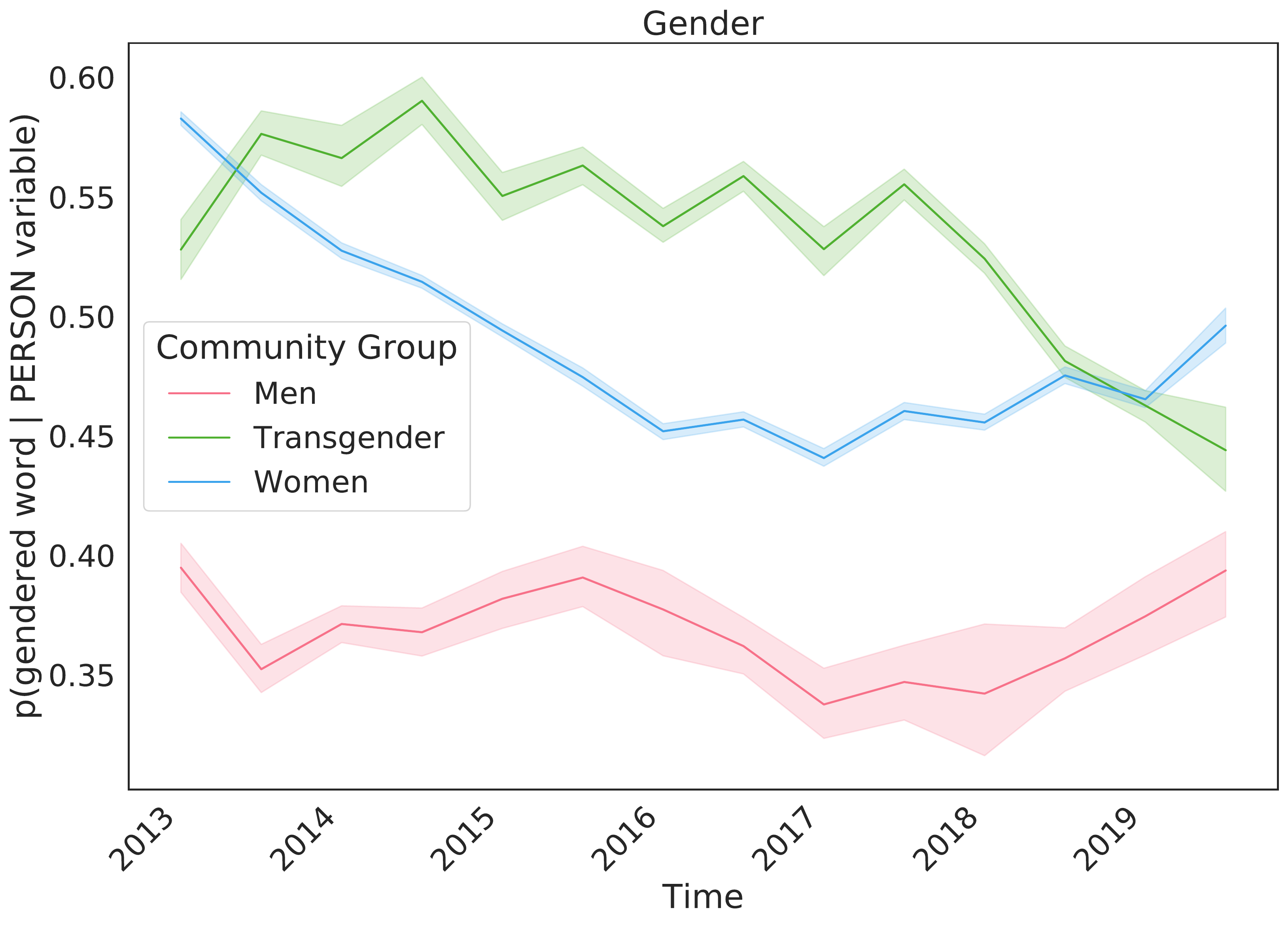}
  
    \caption{ Changes in the probabilities of using gendered \pers terms for Gender and Sexuality-centric communities.
    } 
    \label{fig:pers-temporal}
\end{figure}

A divide in gendered marker usage exists between the sexuality communities. LGBT+ communities ($r$=0.16, $p$<0.01) increase in their use of gendered forms of address compared to their heterosexual ($r$=-0.11, $p$<0.01) counterparts, who gradually use fewer gendered terms. Plots of these trends are shown in \fref{fig:pers-temporal}. %
Although most references are still gendered, these results point to an overall-increased social awareness of the traditional male-norm and shift towards more inclusive gender-neutral language.

\section{Gender Broadening of \textit{dude}}

The previous two studies focused specifically on the analysis of linguistic variables. However, among the terms in both variables, \textit{dude} stands out as a unique address term where a prior study of just that term suggests its usage alone could also reveal changes in attitudes \cite{kiesling2004dude}. Specifically, \textit{dude} can express solidarity with the referents and is occasionally used within female-female interactions, indicating the term is not exclusively a male referent.  Thus, in this third study, we test an additional theory-based term-specific hypothesis that \textit{dude} could undergo semantic widening \cite{bloomfield1945language,blank1999new} where it gradually loses its gender marking and becomes a gender-neutral term that is used to convey solidarity.
Here, we build a computational model to test for gender broadening in \textit{dude} by measuring its relative associations with male and female genders.

\myparagraph{Methods} 
To test for a shift in the gender marking of \textit{dude}, we follow recent methods for bias testing in word embeddings  \cite{caliskan2017semantics,garg2018word,kozlowski2019geometry} and compare the word vector for \textit{dude} with sets of reference vectors that act as semantic poles for measuring its association with male and female genders. %
We use the two datasets from \citet{caliskan2017semantics} that consist of (i) two sets of male and female reference terms, e.g., ``man''  and (ii) two sets of male and female names, which were used to simulate implicit association tests in word semantics. Bias towards one pole (e.g., femininity) is shown by having a higher mean cosine similarity with one set in a pair. 
We further verify the lack of significant synchronic shifts of these anchor words using a Procrustes alignment between sequential year vector spaces. Full details are in supplemental material.

Separate word2vec models \cite{mikolov2013distributed} are trained for each year in our Reddit dataset. Within each year, we compute the mean cosine similarity of \textit{dude} with each word in a set and measure the difference between \textit{dude} and the male and female sets to estimate its gender-association over time.
Similarities are computed over five separate runs on different splits of the aggregate data and then bootstrapped to estimate 95\% confidence intervals. %
For Reddit, word2vec models are trained on a uniform sampling of 10\% of all comments (unfiltered) posted in the first six months of every year, totalling $\sim$8B tokens. %
Here, we calculate Pearson's correlation and perform statistical significance testing over bootstrapped mean probabilities across yearly intervals.

\begin{figure}[t]
    \centering
    \includegraphics[width=0.48\textwidth]{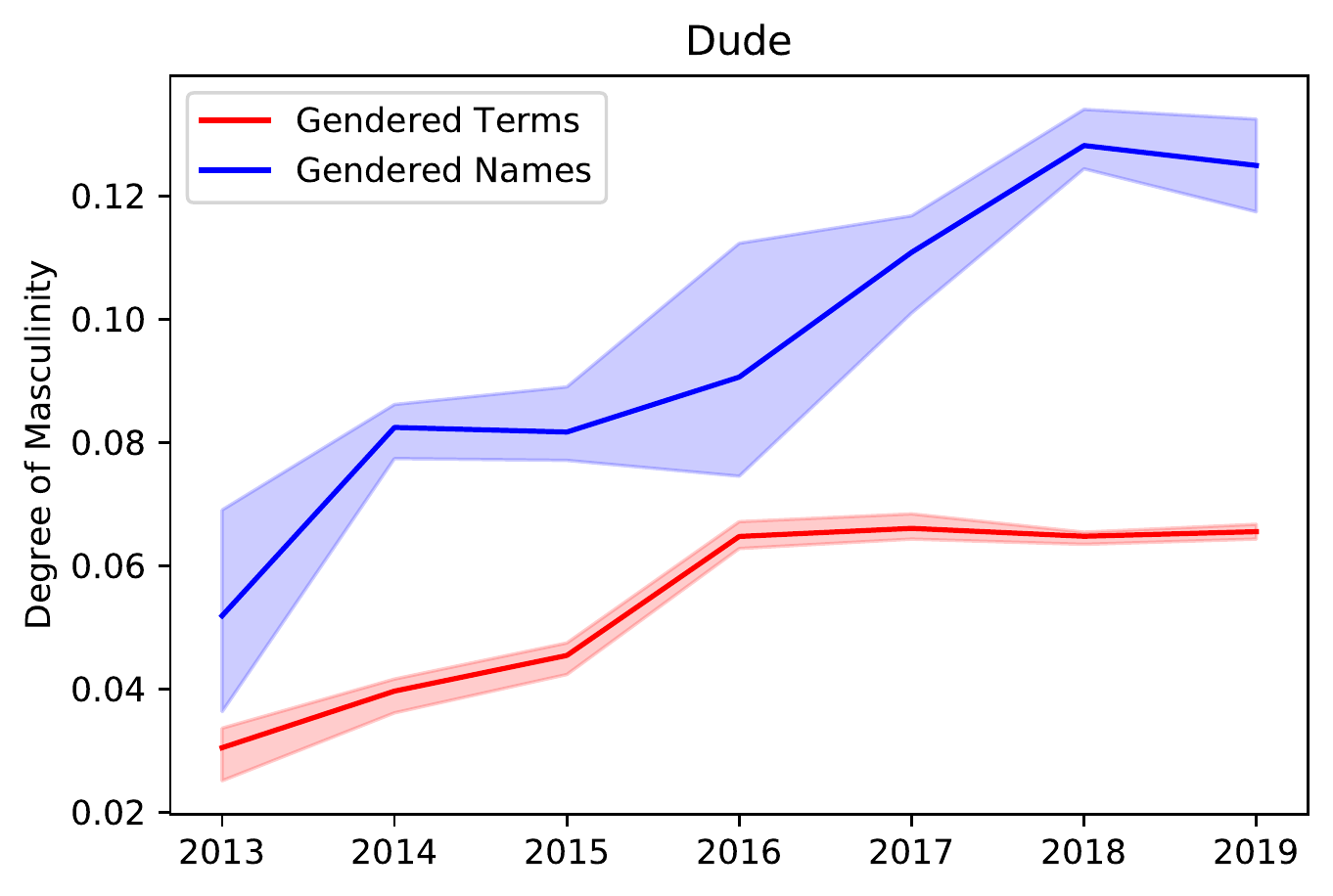}
    \caption{The gender association for \textit{dude} over time reveals semantic narrowing towards an exclusively-masculine use.}
    \label{fig:dude}
\end{figure}

\myparagraph{Results}
The male-gender association for \textit{dude} increases over time for both terms ($r$=0.908, p$<$0.01) and names ($r$=0.962, p$<$0.01) respectively, as shown in \fref{fig:dude}.
This result indicates that \textit{dude} is undergoing a semantic \textit{narrowing}, rather than widening, and increasingly is only used to refer to male referents. 
We view this result as pointing to a general trend towards unambiguous gender markings; whereas prior to the push for gender-neutral English, \textit{dude} may have been widened colloquially, given the increased focus on using ungendered \pers referents, \textit{dude} has narrowed to primarily be used exclusively for male referents.
Further, this view is made with the observation that usage of the word \textit{dude} has increased over time, disallowing a possible explanation that the term's meaning evolved for use in a small semantic niche due a decrease in frequency of use. Together with the results of the \pers variable, these two studies show a marked shift in the linguistic choices marking gender, suggesting broader changes in attitudes about gender.

\section{Discussion}

Sociolinguistic research has consistently shown how subtle variation in language is reflective of attitudes and identity. Our work similarly finds changes in how references to significant others or to indefinite people or groups mirror broader changes in society. However, our study is built on aggregate analyses of social media, which warrants a discussion of potential limitations and caveats, as well as future work.

\myparagraph{Confounding Variables} Compositional demographic changes in the subreddits we study may cause changes in the language use of these communities of practice. As marginalized groups gain increased social acceptance, they may more actively contribute to public forums like Twitter and Reddit. As a result, the observed linguistic changes are also a possibility for a diffusion of linguistic norms that are independent of attitude shifts. Nonetheless, our focus on studying language use in specific communities of practice from the perspective of potential attitude shifts shows that the observed discourse has changed, even if the underlying mechanisms behind that change (changing attitudes or changing group composition) remain to be precisely quantified.

\myparagraph{Demographic Estimates} 
Our study relies on demographic estimates, particularly from using geocoding to infer census-based estimates of persons. However, the American Community Survey Census itself also possesses a degree of bias from a participatory perspective \cite{spielman2014patterns}. Nevertheless, the ACS remains the broadest coverage survey for linking census tracts to demographics.

Further, the composition of users on social media are largely younger and male-dominated (on Reddit, in particular) compared to that of the general population \cite{barthel2016reddit}. Our study has focused on language use in particular online populations whose composition may not reflect Reddit as a whole. While tens of millions of American individuals use these platforms, their participation likely selects a subset of the population whose views do not necessarily generalize to the entire American populace. As a result, future work could test methods (or add additional platforms) to poststratify the analyzed segments of the population to increase representativeness.

\myparagraph{Causality} 
Our study does not make explicit casual claims around factors that may have caused changes in social attitudes---i.e., claims that specific changes in attitude cause this language change. While our work shows evidence of linguistic and attitudinal changes correlated with known policy and legislation changes, like the quasi-causal results estimating the effect of a passage of a Marriage Equality Act on linguistic choices for persons  within that state, we are not arguing these alone explain the change. Our work in no way seeks to diminish the active efforts of folks and social movements that continue, today and that have for decades, striven to advocate for the rights of, change the biased social perceptions towards, and champion the values of equality of traditionally marginalized populations and their lived experiences.
One possibility for moving closer to truly causal studies is through direct participatory work or using causal inference techniques  \cite{feder2021causal} to examine how attitudes influence the word selection or how reading particular uses influence the person's attitude or interpretation of the passage.

\section{Conclusion}

Linguistic choices reflect underlying attitudes about what is being said. To study attitudes about sexuality and gender equality, we identify lexical variables that reflect these attitudes and computationally study these choices using a massive demographically-labeled corpus of 87M English messages from Twitter and Reddit. Our results show that language use has indeed shifted and points to increasing acceptance of non-heterosexual norms towards inclusive, gender-neutral language. Through our demographic analysis, we point to key segments of the population driving these changes.  Further, through a quasi-causal analysis, we show that passage of Marriage Equality Acts in different US states increases the use of gendered spousal references, rather than gender-neutral terms.
While our work does not identify all the underlying causes behind these changes, the results point to where future work could look to identify the structural and social mechanisms behind change and also show how future computational studies can use sociolinguistic variables to tease out demographically-associated attitudes.
Data and code are made available at \url{https://github.com/davidjurgens/sociolinguistic-attitudes}. %

\section{Ethical Considerations}

\myparagraph{Identity Affiliation} 
In studying attitudes, our work aims to characterize attitudes for a particular segment of the population in aggregate, not at the individual level. In doing so, we specifically avoid making strong inferences around a particular actor on social media, e.g., making claims of an individual's gender or sexuality, by only examining behaviors within communities associated with identities. Further, though communities have formed on Reddit around particular identities that are associated with known sociolinguistic variation, participation in these communities does not correspond to direct self-affiliation with these identities. Our methods are instead designed to identify identity-associations, treating participation in these communities as only implicit signals of affiliation with these identities; aiming to identify linguistic communities of practice whose styles may differ in discourse. We also note that in a small set of communities, users can self-select ``flairs'' that explicitly signal an affiliation of some types, e.g., \textit{basketball team fan}, but these are not wide-spread and are often limited only to predefined choice options. In particular, we note the challenges present in treating gender \cite{larson-2017-gender} and sexuality as variables of study, especially in attempts to characterize populations with a faithful regard to gender fluidity. While there is some risk in increasing publicity to communities associated with marginalized identity, we have focused only on larger,  more well-known communities and  avoid ascribing any content to a particular individual.

\myparagraph{External Validity} In particular, we note the concerns by \citet{olteanu2019social} detailing how, among others, there exists (1) self-selecting population biases in social media platforms, (2) behavioral biases regarding user platform content and activities, and (3) content production biases that may vary across demographic groups. Recognizing these biases, we have aimed instead to identify and specifically quantify the differences present between linguistic communities of practice whose styles may differ in online discourse on the platforms we study. While work external to social media has often provided support for our observations, we highlight and point to areas of discrepancies in our findings and encourage future work to further examine these phenomena in non-social media contexts.

\section*{Acknowledgements}
We thank Julia Mendelsohn, Jiaxin Pei, and Jian Zhu for their helpful discussions, Jack Grieve and Dirk Hovy for an initial discussion on this idea at the Lorentz Center (and Nanna Hilton, Dirk Hovy, Dong Nguyen, and Christoph Purschke for organizing that fantastic workshop on Computational Sociolinguistics, which planted the seed that led to this work!), and to the students in the SI 710 PhD seminar on Computational Sociolinguistics for their discussions on this idea. This material is based upon work supported by the National Science Foundation under Grant No.~1850221.

\bibliography{emnlp2020}
\bibliographystyle{acl_natbib}

\clearpage
\newpage

\appendix

\section{Variants} \label{variants}

This section describes the variants considered in the \so and \pers variables, as well as the contextual controls used to filter for them in compiling our datasets.

Variants were selected through a multi-step manual process. We first selected standard terms used in the literature for each variable, e.g., ``husband'' and ``wife.'' We then identified all synonymous terms using multiple thesauri. Finally, a sanity check was done to add any slang or abbreviated versions present in social media using a word-vector-based search and also checking for any common terms used in our patterns that would match. For simplicity, some rare variants were left out; these terms were typically misspellings or word elongations (e.g., ``wiiiiife''). Tables \ref{so-counts} and \ref{pers-counts} show the terms and their valid total counts filtered under contextual controls as they appear on both social media platforms; Table \ref{tab:contexts} shows example extracted uses matching our patterns.

\myparagraph{\so contextual control} All \texttt{PRP\$} usages as tagged by \texttt{NLTK}.

\myparagraph{\pers contextual control} a, an, some, any, both, either, neither, each, every, another, many, most, enough, other, if, when

\begin{table}[h]
    \centering
    \begin{tabular}{r|r}
        \toprule
        \multicolumn{2}{c}{\so} \\
        \midrule
        girlfriend & 24773413 \\
        boyfriend & 33630033 \\
        husband & 16145125 \\
        wife & 2756715 \\
        gf & 7675605 \\
        bf & 8331346 \\
        sweetheart & 1567771 \\
        lover & 2974149 \\
        soulmate & 1215384 \\
        spouse & 1314947 \\
        fiancé & 914207 \\
        fiancée & 311444 \\
        wifey & 450748 \\
        honey & 1065346 \\
        hubby & 1122853 \\
        groom & 74352 \\
        mistress & 458471 \\
        bride & 393800 \\
        darling & 912895 \\
        babe & 4398051 \\
        hon & 37612 \\
        bae & 10455250 \\
        missis & 3492 \\
        partner & 8923329 \\
        \bottomrule
    \end{tabular}
    \caption{Terms and extracted valid total counts of the \so variable.}
    \label{so-counts}
\end{table}

\begin{table}[h]
    \centering
    \begin{tabular}{r|r}
        \toprule
        \multicolumn{2}{c}{\pers} \\
        \midrule
        guy & 45969263 \\
        man & 66342425 \\
        bro & 6800241 \\
        dude & 10705062 \\
        dudette & 5771 \\
        mate & 2175974 \\
        fam & 2160457 \\
        buddy & 3958381 \\
        dawg & 1280205 \\
        pal & 924819 \\
        homie & 52807 \\
        comrade & 93506 \\
        fella & 239641 \\
        girl & 92486478 \\
        person & 42655533 \\
        individual & 4214283 \\
        dawgs & 279870 \\
        dudes & 1770887 \\
        peeps & 566616 \\
        folks & 2292692 \\
        dudettes & 409 \\
        persons & 1072745 \\
        people & 170813192 \\
        bros & 1436324 \\
        guys & 10740018 \\
        girls & 24508620 \\
        comrades & 46582 \\
        buddies & 824542 \\
        pals & 590797 \\
        mates & 1446053 \\
        homies & 1719516 \\
        \bottomrule
    \end{tabular}
    \caption{Terms and extracted valid total counts of the \pers variable.}
    \label{pers-counts}
\end{table}

\vspace{3mm}

\begin{table}[htb]
  \centering
  \begin{tabular}{l} 
  \toprule
  ...I agree. \textbf{A woman} should not be shamed.\\ But \textbf{a man}... \\
  ...\textbf{Many people} have a false sense of \\righteousness. They may...\\
  ...\textbf{if dudes} want to attend a concert,\\they should totally...\\
  ...Long shot, I know, but \textbf{a fella} can dream...\\
  ...Hopefully, \textbf{my husband} will be recovered \\from his cold enough tomorrow to...\\
  ...But this is also what I'm looking\\for in \textbf{my partner}...\\
  ...as \textbf{her lover}. I get to see \textbf{my lover}...\\
  \bottomrule
  \end{tabular}
  \caption{Example contexts. Filtered-for phrases are highlighted in \textbf{bold}. Partial contexts are shown to preserve user anonymity.}
  \label{tab:contexts}
\end{table}

\section{Reddit Communities} \label{reddit_communities}

This section describes the specific subreddits under each of our community-group categorizations for Reddit.

Similar to \citet{hamilton2017loyalty}, Reddit communities were selected and curated in a multi-step process. First, we identified user aggregated lists of themed subreddits; these lists contain subreddits organized by primary theme, e.g., \url{https://www.reddit.com/r/ListOfSubreddits/wiki/listofsubreddits}, as categorized by user opinion in aggregate and only included if they exceeded community-sourced minimum subscriber counts. For each, we then collected related subreddits linked with each community group categorization. Finally, we manually reviewed each subreddit to ensure that its primary theme would mostly fit a particular social group.

\subsection{Politics}

\myparagraph{Right-Leaning} askaconservative, benshapiro, conservative, conservatives, conservativelounge, conservatives\_only, cringeanarchy, jordanpeterson, latestagesocialism, louderwithcrowder, newpatriotism, metacanada, paleoconservative, republican, rightwinglgbt, shitpoliticssays, the\_donald, thenewright, tuesday, walkaway

\myparagraph{Left-Leaning} againsthatesubreddits, accidentallycommunist, anarchafeminism, anarchism, anarchocommunism, anarchosyndicalism, anarchy101, ani\_communism, antiwork, antifascistsofreddit, antifastonetoss, askaliberal, bannedfromthe\_donald, beto2020, bluemidterm2018, breadtube, centerleftpolitics, circlebroke, circlebroke2, chapotraphouse, chapotraphouse2, chomsky, communism, communism101, completeanarchy, dankleft, debateacommunist, debateanarchism, debatecommunism, democrat, democraticsocialism, demsocialists, elizabethwarren, esist, enlightendedcentrism, enoughtrumpspam, enoughlibertarianspam, fragilewhiteredditor, fuckthealtright, fullcommunism, greenparty, impeach\_trump, ironfrontusa, iww, keep\_track, latestagecapitalism, leftwithoutedge, liberal, libertarianleft, libertariansocialism, marchagainsttrump, moretankiechapo, neoliberal, ndp, onguardforthee, ourpresident, pete\_buttigieg, progressive, politics, political\_revolution, politicalhumor, pragerurine, presidentialracememes, russialago, sandersforpresident, selfawarewolves, shitliberalssay, shitthe\_donaldsays, socialdemocracy, socialism, socialism\_101, socialistra, stupidpol, the\_mueller, threearrows, toiletpaperusa, topmindsofreddit, tulsi, voteblue, wayofthebern, yangforpresidenthq

\subsection{Religion}

\myparagraph{General} religion, religioninamerica, faith, philosophyofreligion, debatereligion, explorereligion, abrahamic, tellusofyourgods, elint

\myparagraph{Christianity} christianity, openchristian, catacombs, orthodoxchristianity, catholicism, reformed, pcusa, radicalchristianity, truechristian, quakers, mormon, communityofchrist, christianbooks, theologyclinic, biblestudy, christianapologetics, prayerrequests

\myparagraph{Islam} islam, islamicstudies, progressive\_islam

\myparagraph{Judaism} judaism, talmud

\myparagraph{Eastern} buddhism, theravada, zen, taoism, hinduism

\myparagraph{Non-believers} atheism, losingfaith, agnosticism, humanism, freethought, godlesswomen, exmuslim, exmormon, atheisthavens, atheistssupport

\subsection{Sexuality}

We note one special case for the selection of Sexuality subreddits. The subreddits associated with the Heterosexual identity do contain content related to LGBT+ community members, e.g., same-sex couples will post in relationship\_advice. These posts are the minority of content. However, this overlap is not likely to cause an issue with our analysis due to the direction of the error. Since LGBT+-focused communities largely feature content exclusive to that community, shifts in language of the Heterosexual-associated communities are due to either changes in the actual heterosexual population in those communities or increased participation of LGBT+ community members, both of which signal increased acceptance and normalization, which is the focus of our analysis.

\myparagraph{LGBT+} lgbt, ainbow, asklgbt, happentobegay, glbt, gay, gayrights, lgb, lgbtqadvancement, lgbtnews, lgbtnospam, lgbtvent, lgbtpoliticsblogs, lgbtsex, queer, potofgold, radicalqueers, gaysians, glbtchicago, lgbtindia, lgbtnyc, lgbtpdx, pinke, queerottawa, uklgbt, worldlgbt, actuallesbians, bisexual, bisexuality, pansexual, q4q, masc4masc, meetlgbt, anarchoqueer, cartoon\_gaiety, gaybros, gayclub, gaygeek, gaygineers, gaymers, gaymersgonemild, gayreads, gaysports, gaytheists, gbltcompsci, happentobegay, lesbients, lgbt\_cartoons, lgbtcirclejerk, lgbtunes, lgbtrees, qpoc, queercore, queerfashionadvice, queercinema, samesexparents, thecloset, tranarchism, transhack, gayyoungold, lgbtolder, lgbtqteens, queeryouth, comingoutsupport, itgetsbetter, lgbtquestions, pflag, asexual, asexuality

\myparagraph{Heterosexual} relationship\_advice, dating, dating\_advice, relationships

\subsection{Gender}

\myparagraph{Transgender} androgynoushotties, asktransgender, crossdressing, drag, queertransmen, rslashtransgenderfaq, tgdisc, transgender, transphobiaproject, transspace

\myparagraph{Men} daddit, malestudies, malefashionadvice, malehairadvice, malelifestyle, men, mrr, oney, xychromosomes

\myparagraph{Women} askwomen, babybumps, entwives, feminisms, feminism, girlgamers, mommit, twoxchromosomes, women, xxfitness

\section{Association with Masculine References} 
\label{word2vec_vectors}

This section describes the specific reference terms used in our study to determine the association of \textit{dude} with masculine references. Reference terms noted here were drawn from the supplemental material in \cite{caliskan2017semantics} as names and words used to study gender. %

\vspace{3mm}

\myparagraph{Male Names} john, paul, mike, kevin, steve, greg, jeff, bill

\myparagraph{Female Names} amy, joan, lisa, sarah, diana, kate, ann, donna

\myparagraph{Male Terms} male, man, boy, brother, he, him, his, son, father, uncle, grandfather

\myparagraph{Female Terms} female, woman, girl, sister, she, her, hers, daughter, mother, aunt, grandmother

\vspace{3mm}

To validate against the possibility of synchronic shifts, we compute cosine similarities for these anchor words following a Procrustes alignment between sequential year vector spaces on our yearly word2vec models trained across samples of all posts and comments on Reddit and Twitter. Shown in Table \ref{tab:procrustes}, a high degree of cosine similarity was present for all anchor words, suggesting no significant synchronic shifts for these anchor words occurred.

\section{Word2Vec Training Details}

This section lists the details and specific hyperparameters used for word2vec \cite{mikolov2013distributed} model training.

\vspace{3mm}

We train all word embeddings in word2vec with \texttt{gensim} \cite{rehurek_lrec}, setting word vector dimensionality to 300 with a continuous bag of words (CBOW) architecture. Training was performed until stable loss convergence, which resulted for all models at around 15 epochs. Other hyperparameters were left unchanged from library defaults. Training was performed on 50 Intel Xeon CPU cores and times ranged from 30 minutes to 2 hours, which varied according to dataset size.

\section{Expanded Results} \label{expanded-results}

\begin{figure*}[!t]
    \centering
    \begin{tabular}{cccc}
    \subfloat[Sexuality]{\includegraphics[width=0.22\textwidth]{imgs/plots-final/partner-prp-all/sexuality.pdf}} & 
    \subfloat[Gender]{\includegraphics[width=0.22\textwidth]{imgs/plots-final/partner-prp-all/gender.pdf}} &  
    \subfloat[Political Leaning]{\includegraphics[width=0.22\textwidth]{imgs/plots-final/partner-prp-all/politics.pdf}} &    
    \subfloat[Religion]{\includegraphics[width=0.22\textwidth]{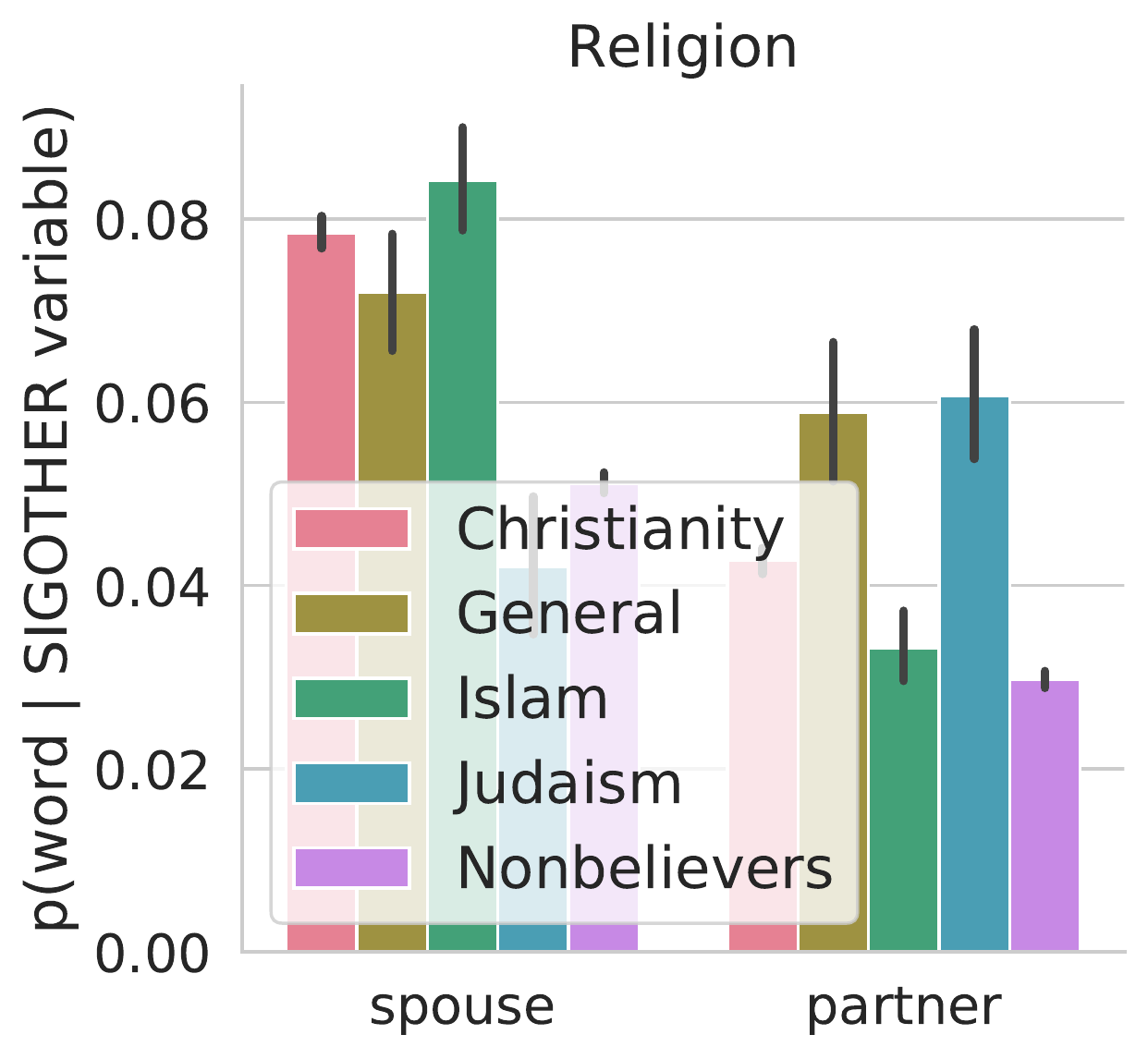}} \\   
    \subfloat[Urban Density]{\includegraphics[width=0.22\textwidth]{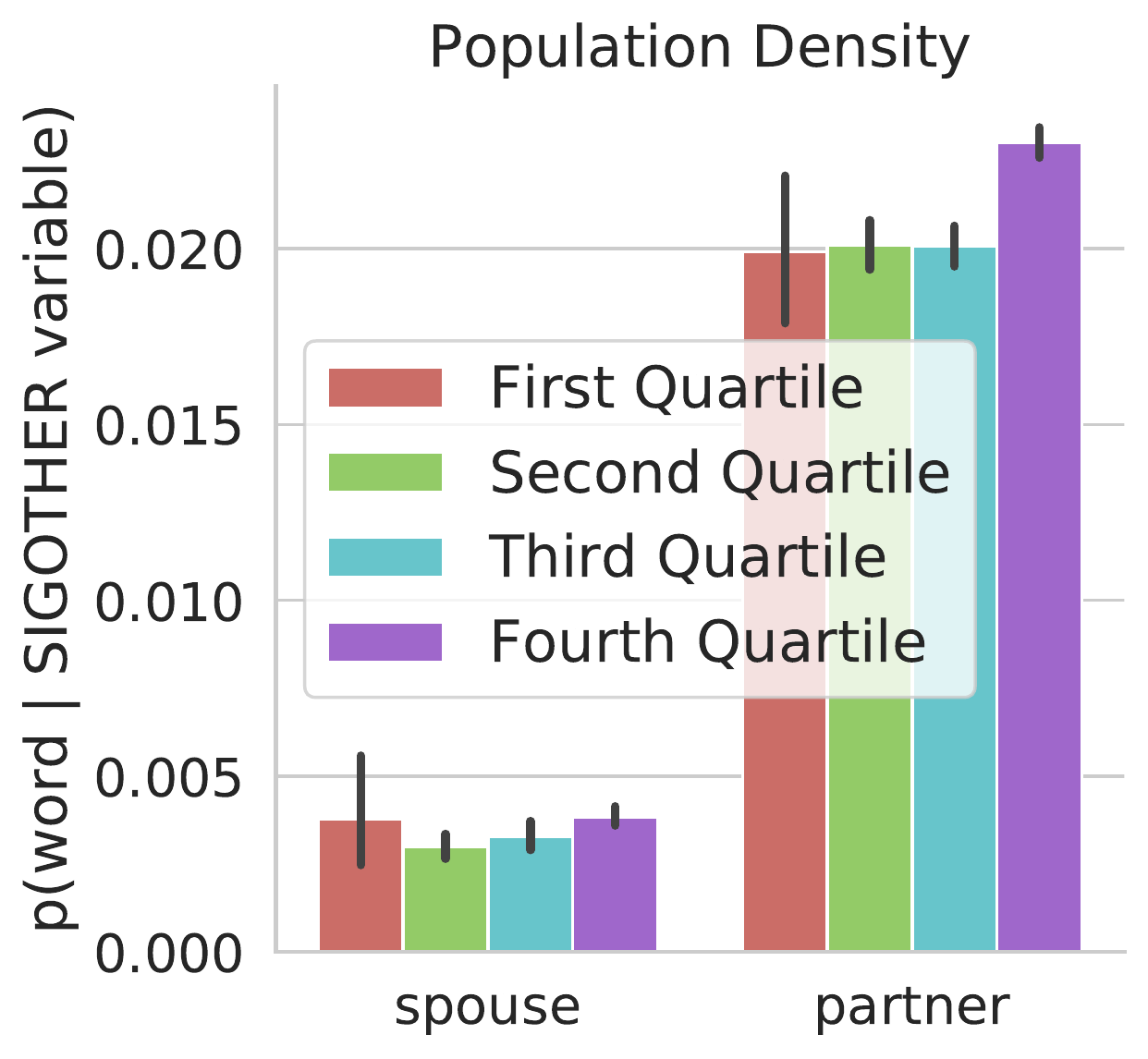}} & 
    \subfloat[Education]{\includegraphics[width=0.22\textwidth]{imgs/plots-final/partner-prp-all-ses/var_percent_with_graduateandhigher_25andolder.pdf}} &   
    \subfloat[Income]{\includegraphics[width=0.22\textwidth]{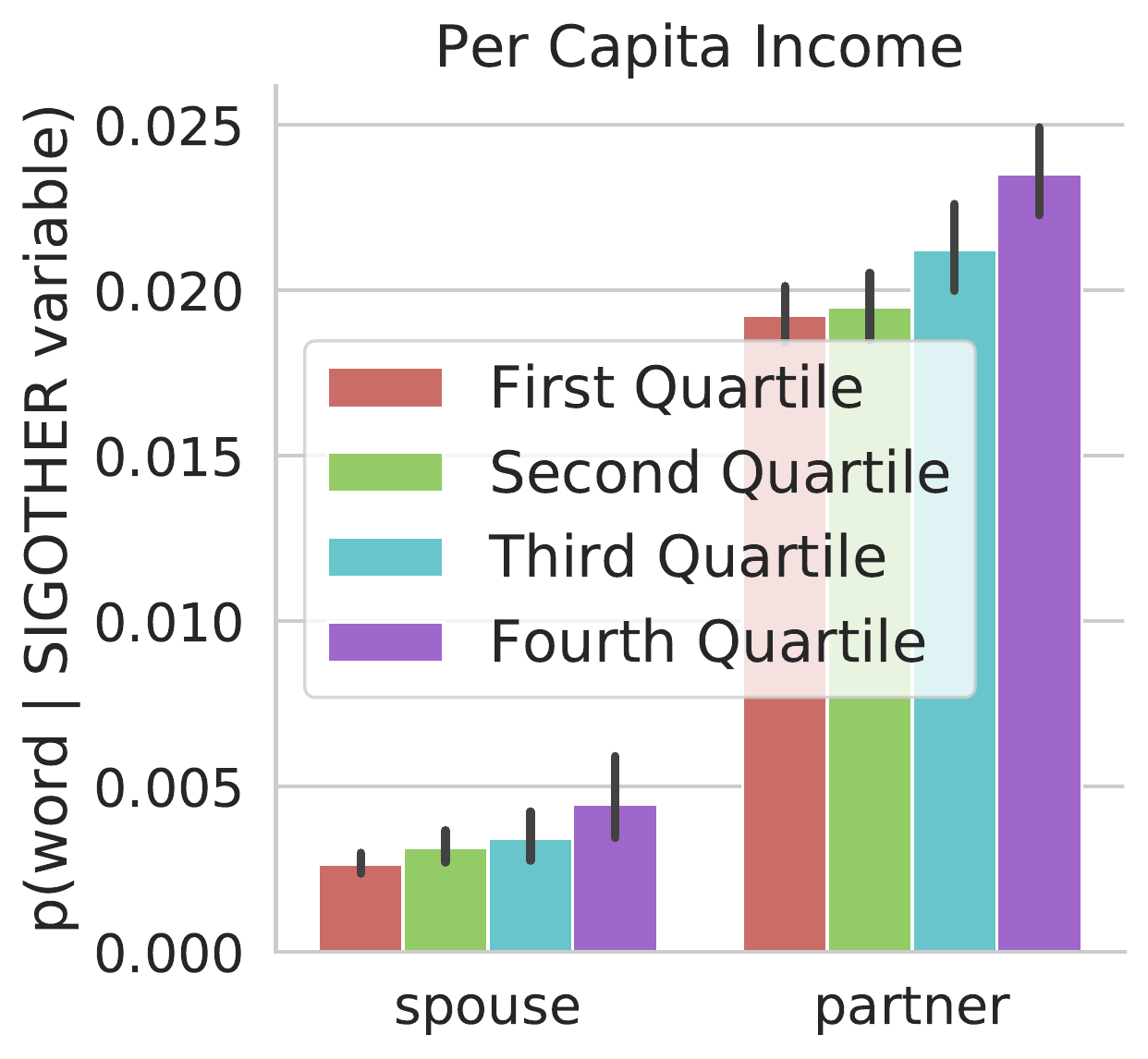}} &
    \subfloat[Inequality]{\includegraphics[width=0.22\textwidth]{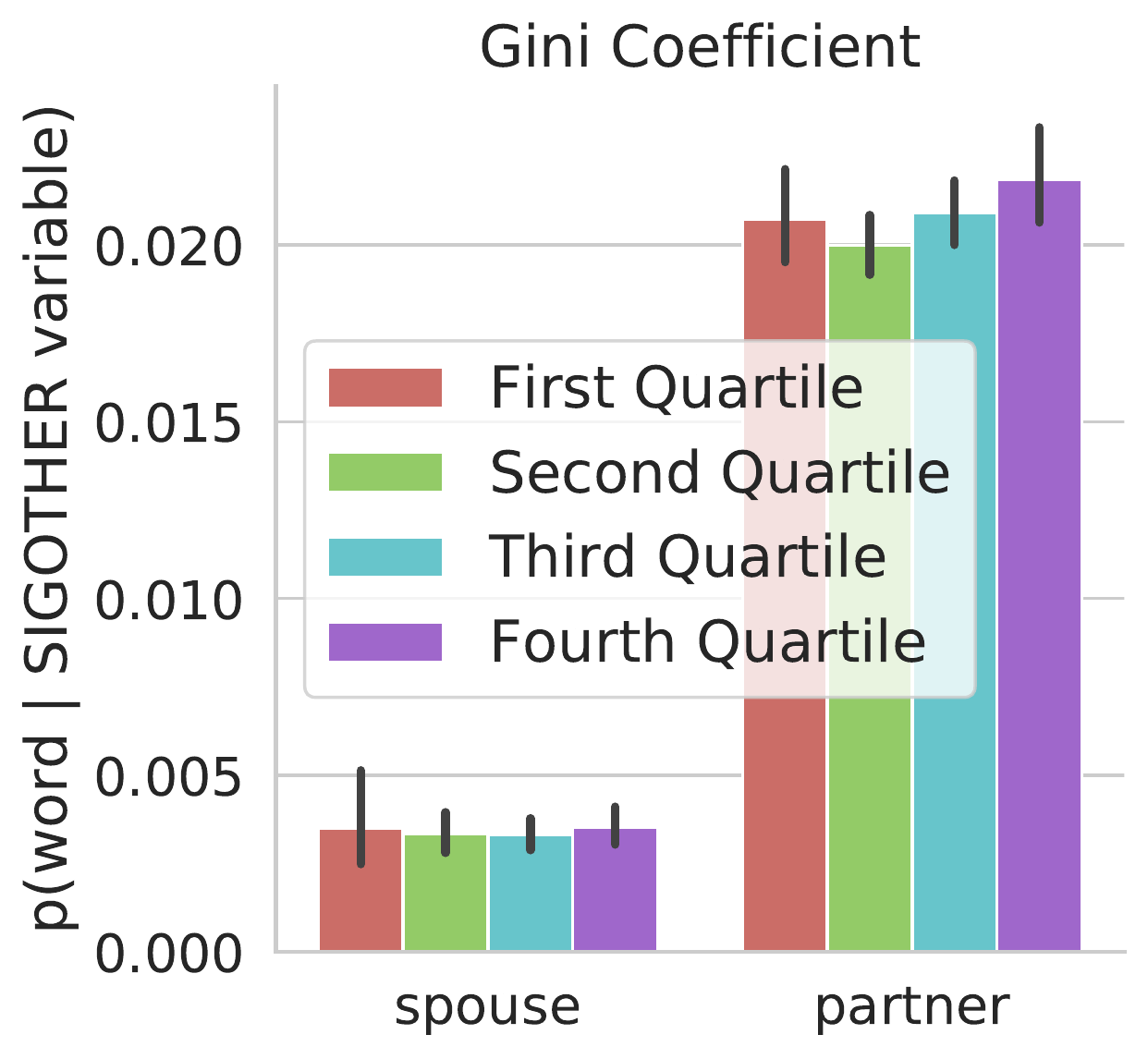}}\\  
    \end{tabular}    

    \caption{  The probabilities of \textit{partner} and \textit{spouse} use in \so within identity-aligned communities. 95\% confidence intervals are shown.  
    }
    \label{who-uses-so-appendix}
\end{figure*}

\begin{figure*}[!t]
    \centering
    \resizebox{\textwidth}{!}{
    \begin{tabular}{cccc}
    \subfloat[Sexuality]{\includegraphics[width=0.24\textwidth]{imgs/plots-final/dude-controlled/sexuality.pdf}} & 
    \subfloat[Gender]{\includegraphics[width=0.24\textwidth]{imgs/plots-final/dude-controlled/gender.pdf}} &
    \subfloat[Political Leaning]{\includegraphics[width=0.24\textwidth]{imgs/plots-final/dude-controlled/politics.pdf}} &    
    \subfloat[Religion]{\includegraphics[width=0.24\textwidth]{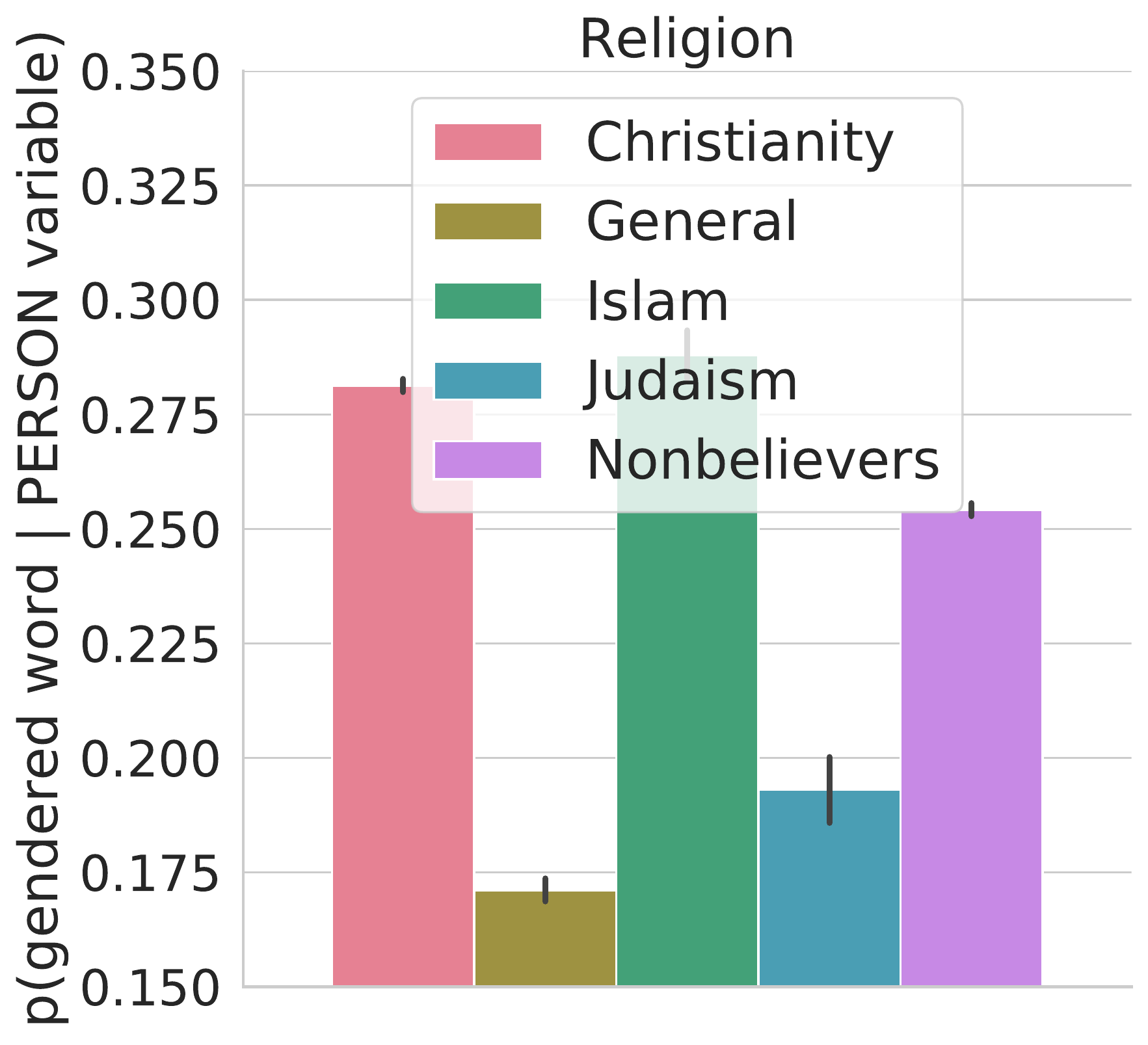}} \\  
    \subfloat[Urban Density]{\includegraphics[width=0.24\textwidth]{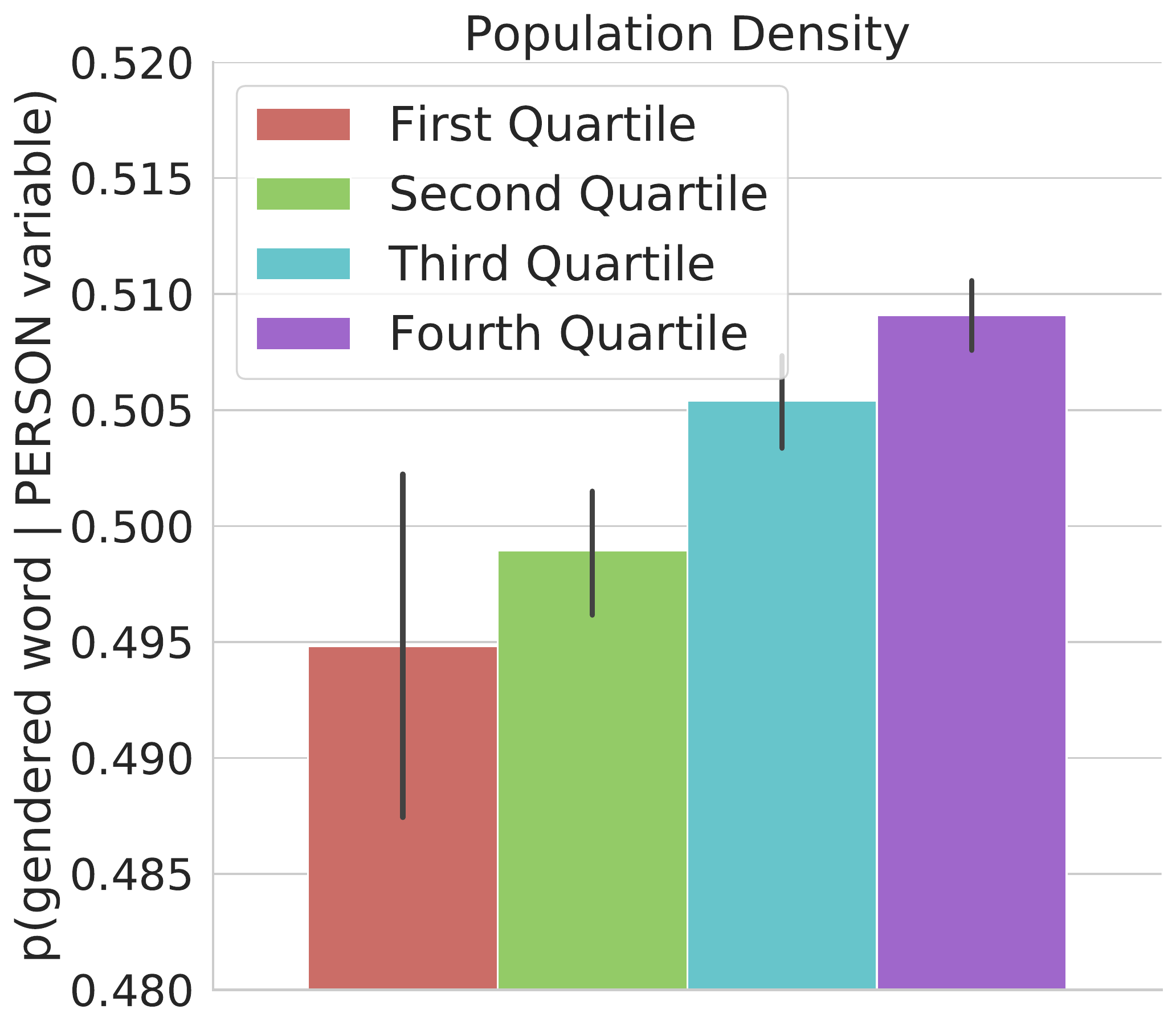}} & 
    \subfloat[Education]{\includegraphics[width=0.24\textwidth]{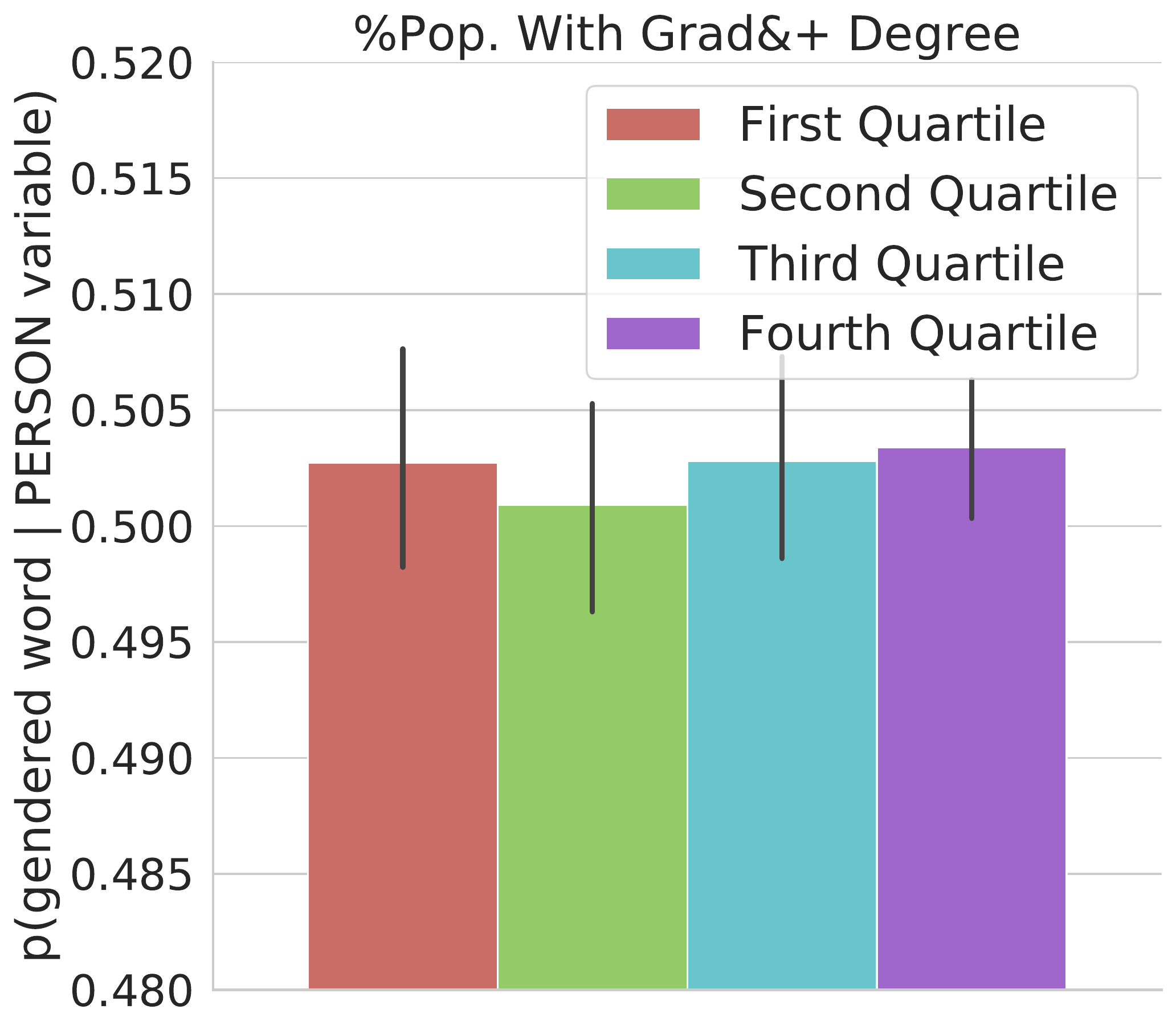}} &
    \subfloat[Income]{\includegraphics[width=0.24\textwidth]{imgs/plots-final/dude-controlled-ses/var_per_capita_income.pdf}} &
    \subfloat[Inequality]{\includegraphics[width=0.24\textwidth]{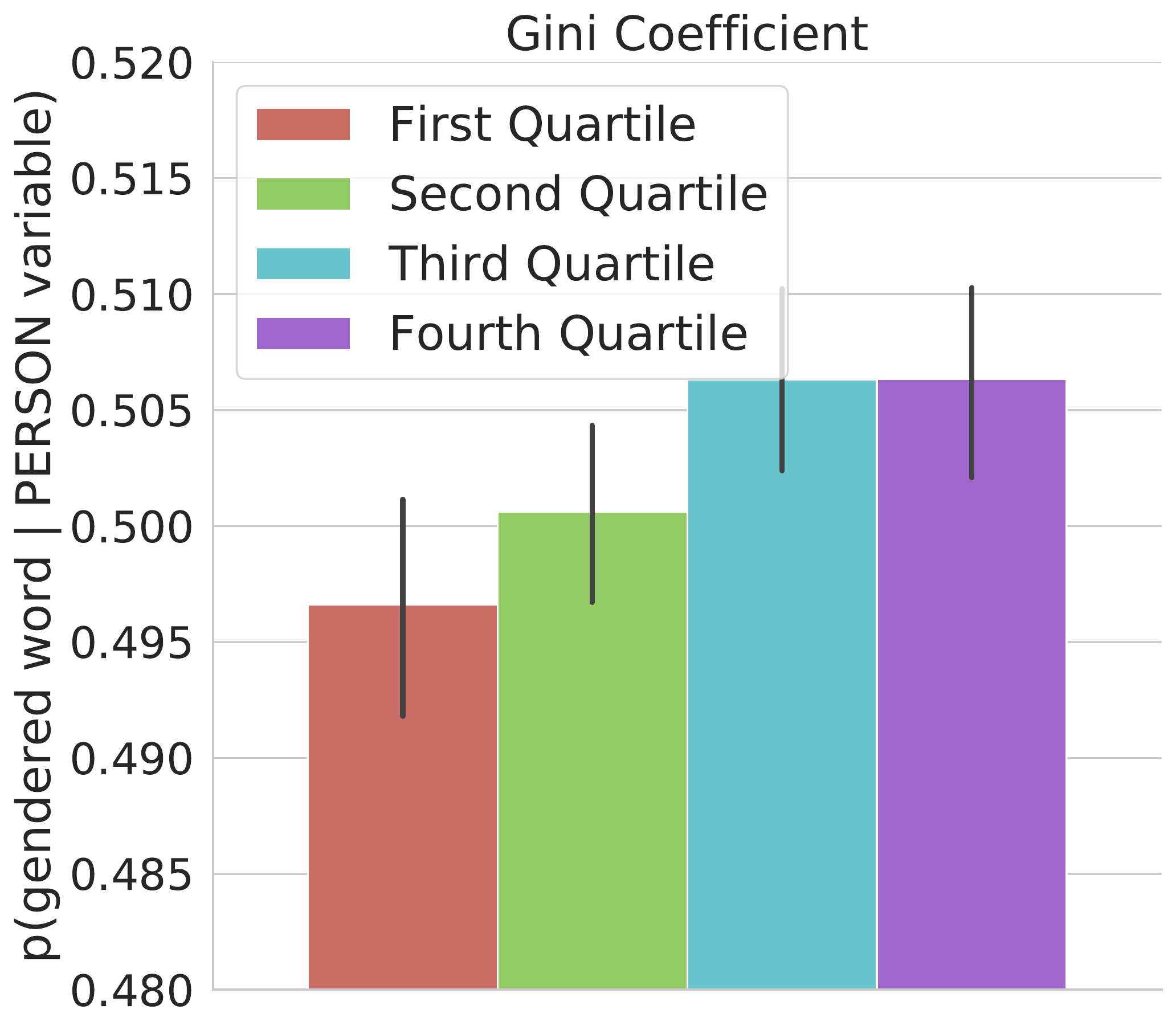}} \\  
    \end{tabular}    
    }
    \caption{ 
    The probabilities of using gendered \pers terms within identity-aligned communities. 95\% confidence intervals are shown.
    }
    \label{who-uses-pers-appendix}
\end{figure*}

\begin{figure*}[!t]
    \centering
    \includegraphics[width=0.32\textwidth]{imgs/plots-final/dude-controlled-time/sexuality.pdf}
    \includegraphics[width=0.32\textwidth]{imgs/plots-final/dude-controlled-time/gender.pdf}
    \includegraphics[width=0.32\textwidth]{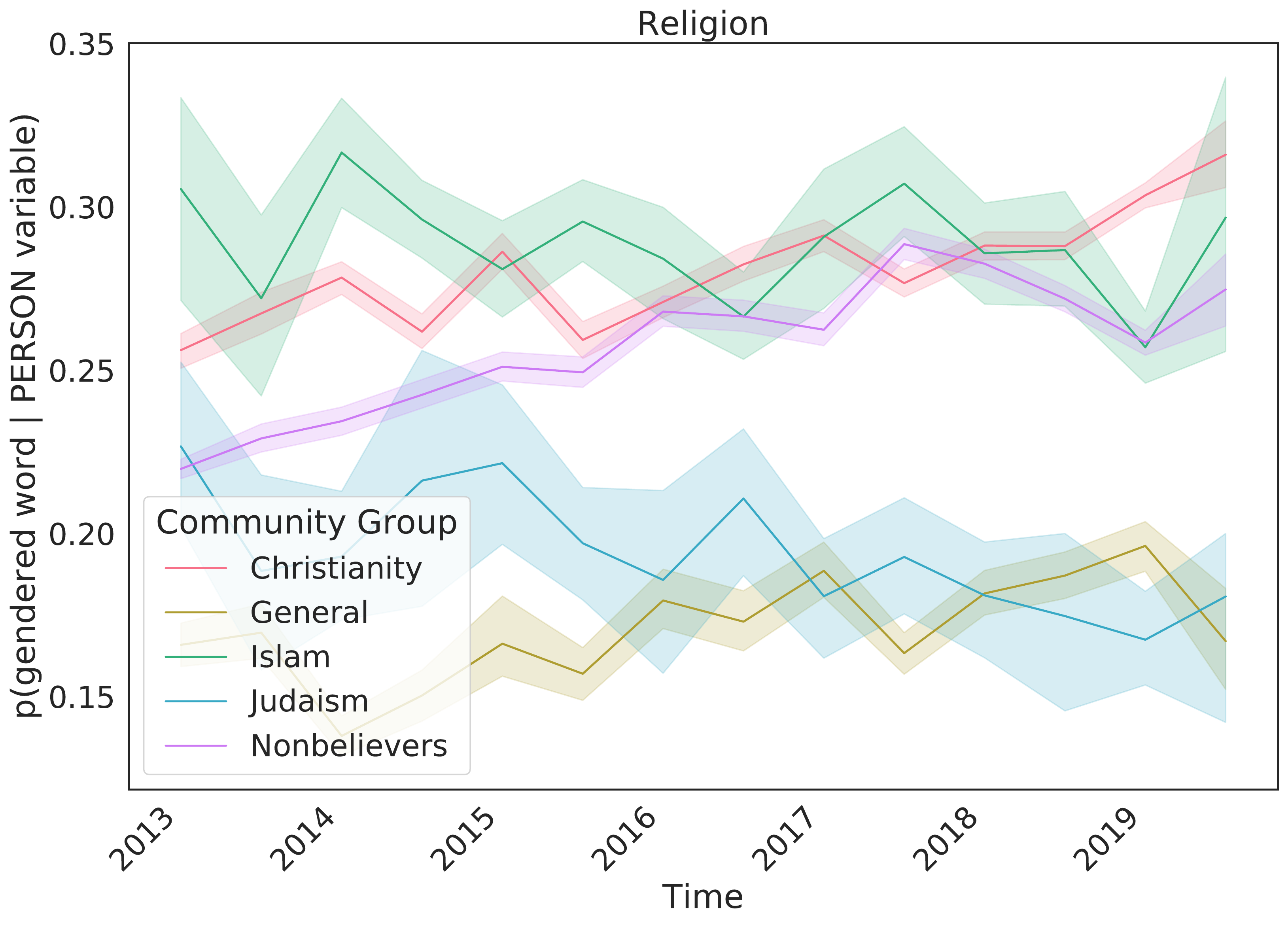}
  
    \caption{ Changes in the probabilities of using gendered \pers terms for Gender, Sexuality, and Religion-centric communities. 95\% confidence intervals are shown.} 
    \label{who-uses-pers-time-appendix}
\end{figure*}

Expanded results for gendered \pers use among identity-centric communities and associations with different socioeconomic status variables are shown in Figure \ref{who-uses-pers-appendix}, with their variations over time shown in Figure \ref{who-uses-pers-time-appendix}. Similar results for \textit{spouse} and \textit{partner} use in \so are shown in Figure \ref{who-uses-so-appendix}.

We additionally include two case studies in the expanded results looking at the use of \pers and \so in specific communities of practice.

\subsection{Religion}

Discussions in religious communities are less likely to use gendered markers of \pers relative to political, gender, and sexuality communities. This mirrors the finding that the recognition of and pushes towards more inclusive language has been prevalent in religious communities \cite{hardesty1987inclusive,cochran2005evangelical}, especially in Judaism \cite{adler1998engendering}, which has seen a multitude of feminist and progressive views \cite{raphael2003female}, while progressive movements within atheist communities are still in the minority \cite{kettell2014divided}.

\subsection{Community-Specific Case Study: Asexuality} \label{asexual-case-study}

\begin{figure}[t]
    \centering
    \begin{tabular}{c}
    \subfloat[Gendered \pers]{\includegraphics[width=0.32\textwidth]{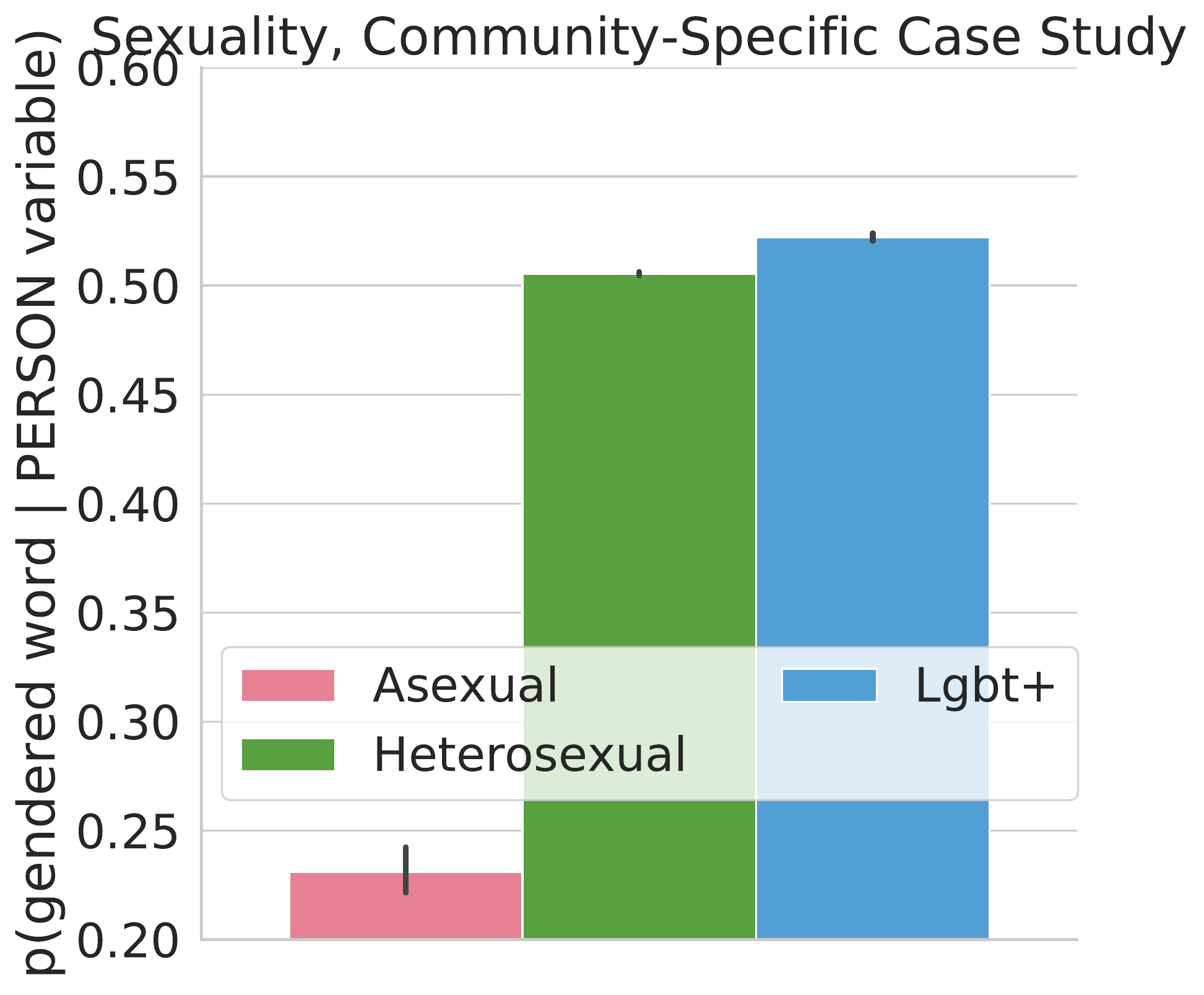}} \\
    \subfloat[partner/spouse \so]{\includegraphics[width=0.32\textwidth]{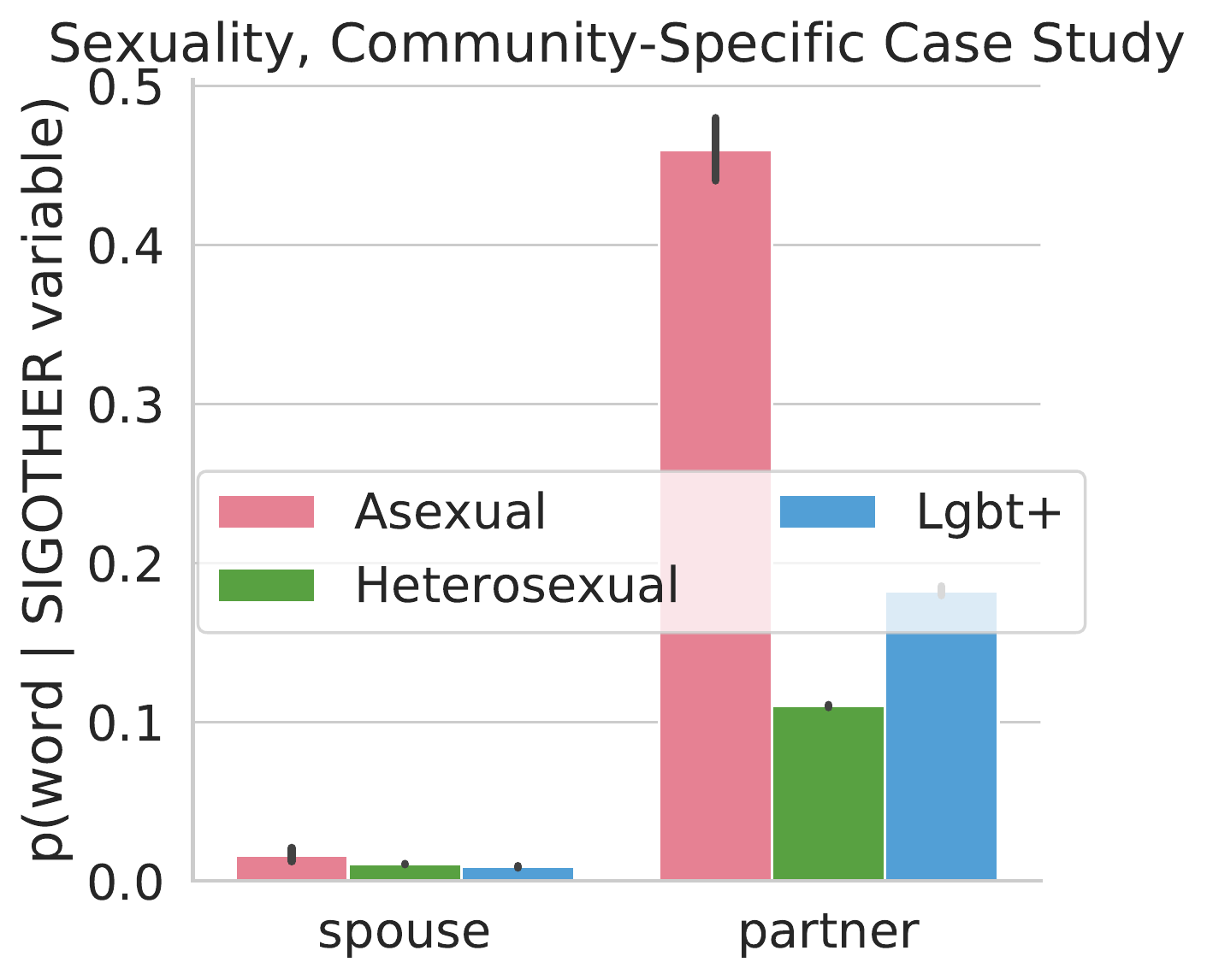}} \\
    \end{tabular}
  
    \caption{The probabilities of gendered \pers use, as well as \textit{partner} and \textit{spouse} use in \so, for Asexual communities relative to their use in the heterosexual and greater LGBT+ communities. 95\% confidence intervals are shown.} 
    \label{asexual-fig}
\end{figure}

Asexuality is a complex self-categorization with asexual sub-identities often referring to relationship preferences and/or an aromantic orientation \cite{bogaert2006toward, prause2007asexuality, macneela2015freedom}. A significant portion of asexual individuals identify with gender-neutral labels over traditional male/female binary categories \cite{brotto2010asexuality}; here, we quantify \pers and \so variable use in subreddits \texttt{r/asexual} and \texttt{r/asexuality} and compare it to the greater LGBT+ as well as heterosexual communities. Results, illustrated in Figure \ref{asexual-fig}, show that discussions in asexual communities are significantly more likely to use gender-neutral markers of both \pers and \so relative to the greater LGBT+ as well as heterosexual communities, mirroring aforementioned self-categorization survey findings.

\subsection{Community-Specific Case Study: Trans-Exclusionary Radical Feminism}

\begin{figure}[t]
    \centering
    \begin{tabular}{c}
    \subfloat[Gendered \pers]{\includegraphics[width=0.32\textwidth]{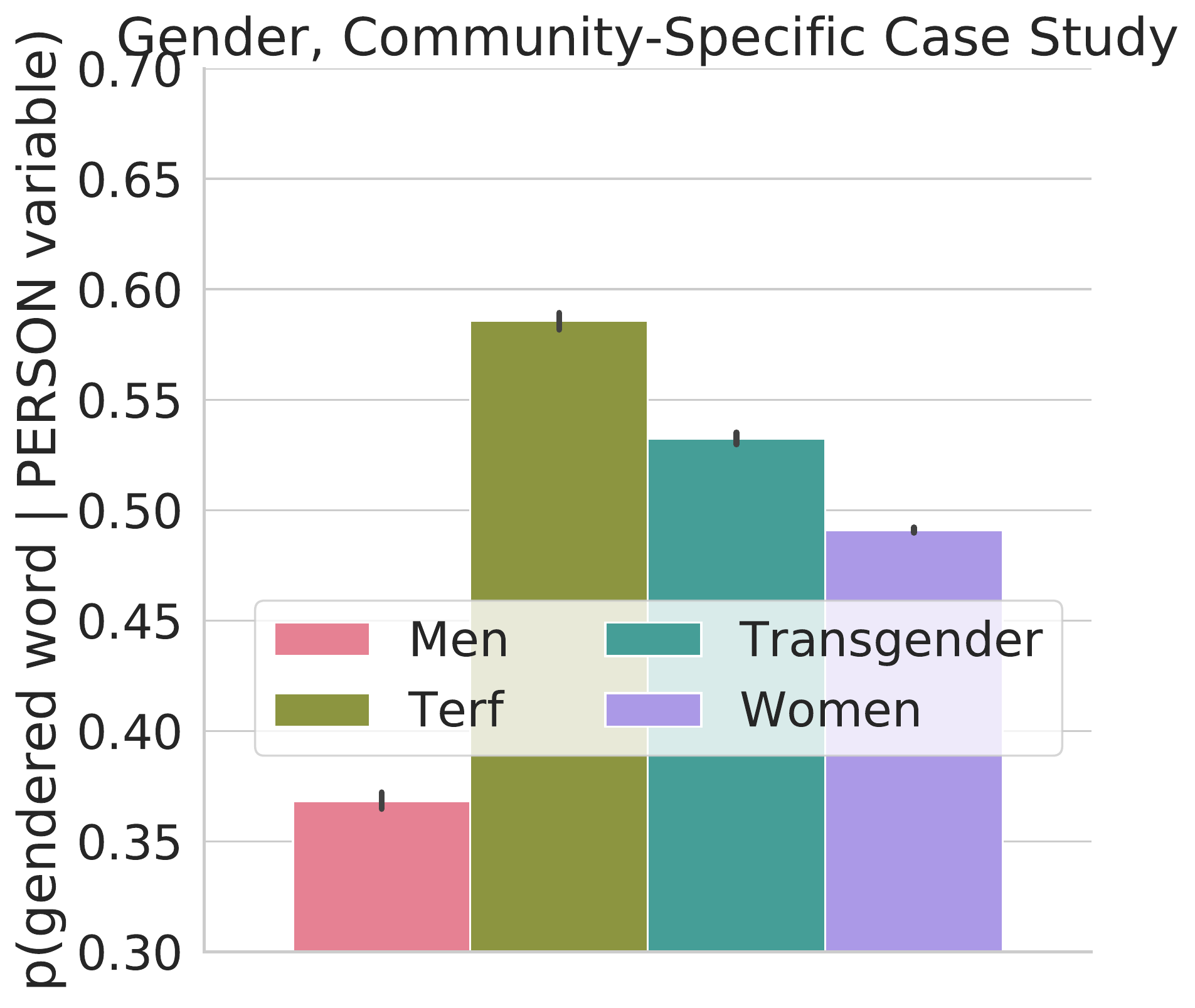}} \\
    \subfloat[partner/spouse \so]{\includegraphics[width=0.32\textwidth]{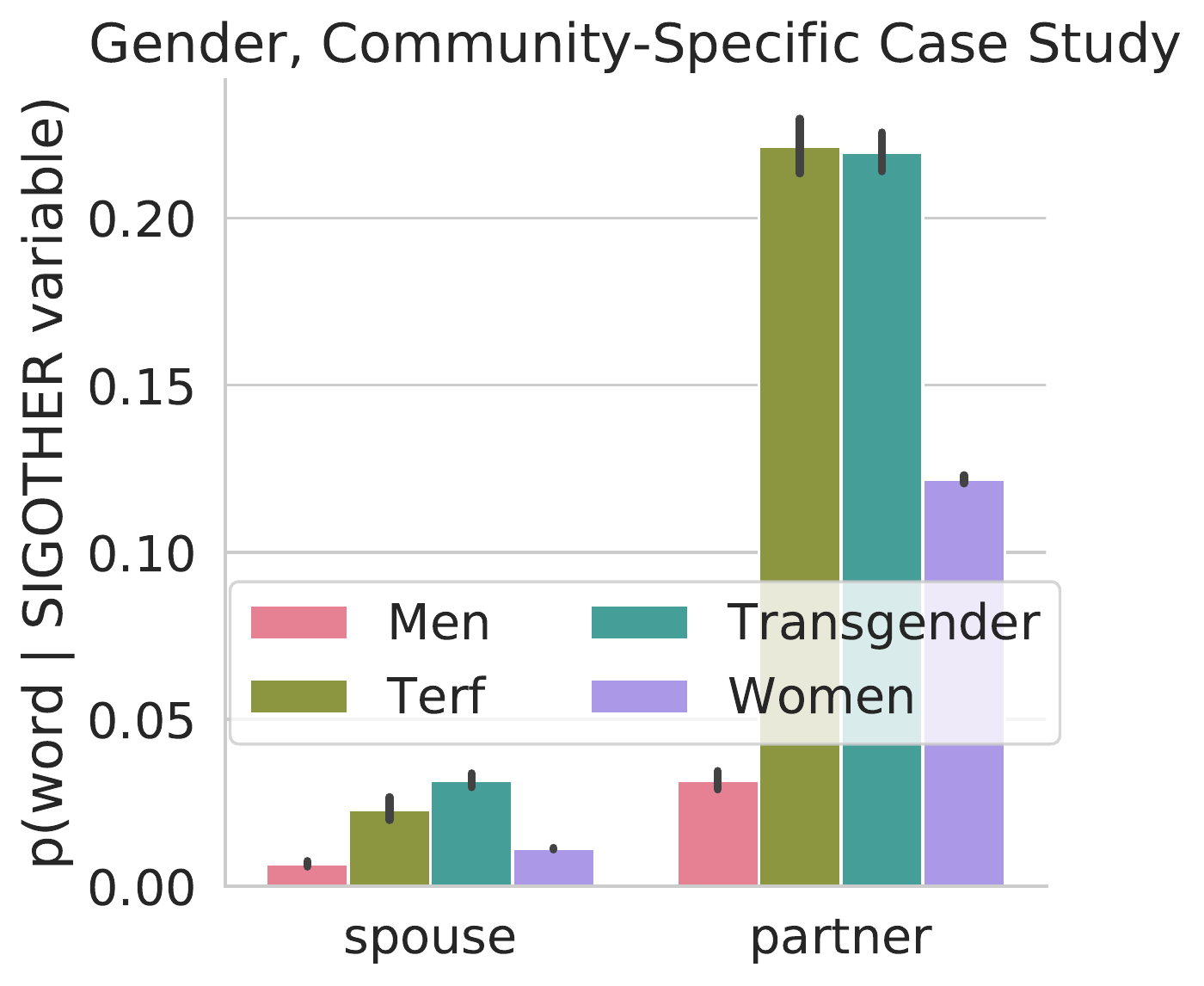}} \\
    \end{tabular}
  
    \caption{The probabilities of gendered \pers use, as well as \textit{partner} and \textit{spouse} use in \so, for TERF communities relative to other communities of practices centered on particular gender identities. 95\% confidence intervals are shown.} 
    \label{terf-fig}
\end{figure}

Trans-exclusionary radical feminists (TERFs) or ``gender critical'' feminists, a self-identified transphobic hate group community, propagate transphobia under the guise of feminism \cite{pearce2020terf}. Following recent interest in the analysis of TERF community online behavior \cite{lu2020computational}, here, we quantify \pers and \so variable use in the \texttt{/r/gendercritical} subreddit---by and large the most prominent\footnote{The subreddit was banned by Reddit in 2020 for ``violating Reddit’s rule against promoting hate''; community posts from 2013-2019 were collected as part of our analysis.} TERF community on Reddit---and compare this community against variable usage in other communities of practice centered on particular gender identities. We hypothesize that the gender-binary viewpoint taken by the TERF community would lead to sharply different language use. %

Results, illustrated in Figure \ref{terf-fig}, show that discussions in the TERF community are more likely to use gendered markers of \pers relative to those in other gender-related communities of practice. This high frequency of gender marking could reflect increased topical content on gender (content-driven) or the groups' stronger focus on highlighting gender identity as salient (attitude-driven). However,  conversations in the TERF communities are also more likely to use gender-neutral markers when referring to \so---which by construction are primarily references to one's own spouse/partner (unless used in a quote, which is rare). The rate of ``partner'' is statistically equivalent to the rate seen in communities of practice focused on transgender issues and identities. This behavior suggests divergent marking of gender: TERF users are more likely to mark gender when referring to \textit{others}, but less likely to mark gender when referring to one's \textit{own} significant other. We view this unexpected result as pointing to an opportunity for future studies of the mechanisms behind this linguistic behavior, as the different practices of gender marking displayed in this community are not seen elsewhere.

\begin{table*}
    \centering
    \begin{tabular}{r|rrrrrr}
        \toprule
        &  \multicolumn{5}{c}{Cosine Similarity, Sequential Years}\\
        Word & 2013\textemdash 2014 & 2014\textemdash 2015 & 2015\textemdash 2016 & 2016\textemdash 2017 & 2017\textemdash 2018 & 2018\textemdash 2019 \\
        \midrule
john & 0.991 & 0.994 & 0.985 & 0.983 & 0.986 & 0.979 \\
paul & 0.985 & 0.991 & 0.984 & 0.965 & 0.967 & 0.964 \\
mike & 0.990 & 0.981 & 0.984 & 0.982 & 0.979 & 0.974 \\
kevin & 0.981 & 0.977 & 0.975 & 0.951 & 0.968 & 0.975 \\
steve & 0.988 & 0.988 & 0.981 & 0.967 & 0.974 & 0.959 \\
greg & 0.955 & 0.882 & 0.879 & 0.899 & 0.868 & 0.895 \\
jeff & 0.981 & 0.983 & 0.971 & 0.970 & 0.950 & 0.952 \\
bill & 0.987 & 0.987 & 0.979 & 0.984 & 0.986 & 0.993 \\
amy & 0.984 & 0.983 & 0.970 & 0.960 & 0.971 & 0.962 \\
joan & 0.982 & 0.966 & 0.968 & 0.958 & 0.972 & 0.975 \\
lisa & 0.977 & 0.977 & 0.948 & 0.972 & 0.966 & 0.959 \\
sarah & 0.986 & 0.980 & 0.987 & 0.969 & 0.967 & 0.953 \\
diana & 0.985 & 0.982 & 0.971 & 0.933 & 0.925 & 0.985 \\
kate & 0.988 & 0.981 & 0.966 & 0.955 & 0.952 & 0.943 \\
ann & 0.984 & 0.976 & 0.982 & 0.986 & 0.978 & 0.977 \\
donna & 0.989 & 0.978 & 0.976 & 0.967 & 0.975 & 0.985 \\
male & 0.998 & 0.994 & 0.995 & 0.994 & 0.994 & 0.994 \\
man & 0.994 & 0.991 & 0.991 & 0.993 & 0.991 & 0.989 \\
boy & 0.991 & 0.994 & 0.993 & 0.990 & 0.992 & 0.986 \\
brother & 0.997 & 0.995 & 0.994 & 0.991 & 0.995 & 0.991 \\
he & 0.998 & 0.997 & 0.998 & 0.996 & 0.997 & 0.994 \\
him & 0.997 & 0.996 & 0.996 & 0.994 & 0.995 & 0.994 \\
his & 0.996 & 0.994 & 0.992 & 0.992 & 0.993 & 0.992 \\
son & 0.996 & 0.996 & 0.992 & 0.989 & 0.990 & 0.990 \\
father & 0.996 & 0.995 & 0.994 & 0.993 & 0.994 & 0.994 \\
uncle & 0.995 & 0.995 & 0.992 & 0.991 & 0.990 & 0.988 \\
grandfather & 0.995 & 0.999 & 0.994 & 0.993 & 0.994 & 0.993 \\
female & 0.997 & 0.994 & 0.993 & 0.992 & 0.991 & 0.991 \\
woman & 0.996 & 0.995 & 0.989 & 0.991 & 0.991 & 0.992 \\
girl & 0.995 & 0.997 & 0.994 & 0.994 & 0.992 & 0.993 \\
sister & 0.996 & 0.996 & 0.992 & 0.994 & 0.994 & 0.994 \\
she & 0.997 & 0.996 & 0.994 & 0.994 & 0.995 & 0.994 \\
her & 0.997 & 0.997 & 0.994 & 0.991 & 0.992 & 0.994 \\
hers & 0.993 & 0.993 & 0.988 & 0.990 & 0.992 & 0.991 \\
daughter & 0.997 & 0.997 & 0.996 & 0.994 & 0.995 & 0.994 \\
mother & 0.994 & 0.996 & 0.995 & 0.995 & 0.995 & 0.994 \\
aunt & 0.997 & 0.994 & 0.994 & 0.996 & 0.997 & 0.994 \\
grandmother & 0.996 & 0.996 & 0.994 & 0.996 & 0.994 & 0.995 \\
        \bottomrule
    \end{tabular}
    \caption{Cosine Similarities for anchor words following a Procrustes alignment between sequential year vector spaces on our yearly word2vec models.}
    \label{tab:procrustes}
\end{table*}

\end{document}